\documentclass[journal]{IEEEtran}
\usepackage{amsmath, amssymb, amsfonts}

\usepackage{booktabs}
\usepackage{multirow}
\usepackage{ifpdf}
\usepackage{placeins}

\usepackage{cite}
\usepackage{stfloats}
\usepackage{afterpage}

\ifCLASSINFOpdf
\usepackage[pdftex]{graphicx}
\else
\fi
	\usepackage{float}   % 支持 [H]
	
	\usepackage{graphicx}
	\usepackage{makecell}
	\usepackage{xcolor}
	\usepackage[colorlinks=true,
	linkcolor=blue,
	citecolor=blue,
	urlcolor=blue]{hyperref}
	
\begin{document}
		\title{TriDifNet: A Unified Three-Task Diffusion Network for End-to-End RGB-T Perception}

		\author{Qianchao He, Jing Liu,~\IEEEmembership{Senior Member,~IEEE,} Yunhao Li and Xiaoqing Tian
			
			\thanks{This work was supported in part by the National Natural Science Foundation of China under Grant
				62173267, Grant 61573276, and Grant 62173266. (\textit{Corresponding author}: Jing Liu.)}
			\thanks{Qianchao He, Jing Liu and Yunhao Li are with the School of Automation Science and Engineering, Faculty of Electronic and Information Engineering, Xi'an Jiaotong University, Xi'an 710049, China (e-mail: banyedisco@\allowbreak 163.com; elelj20080730@\allowbreak gmail.com; lyhxjstu@\allowbreak stu.xjtu.edu.cn).}
			\thanks{Xiaoqing Tian is with the School of Electrical and Control Engineering, Xi'an University of Science and Technology, Xi'an 710054, China (e-mail: tianxiaoqing2017@\allowbreak xust.edu.cn).}
			% IEEE preprint notice (IEEE Section 3.2); do not replace with published-paper copyright.
			\thanks{This work has been submitted to the IEEE for possible publication. Copyright may be transferred without notice, after which this version may no longer be accessible.}}

		\markboth{IEEE TRANSACTIONS ON CIRCUITS AND SYSTEMS FOR VIDEO TECHNOLOGY}%
		{He \MakeLowercase{\textit{et al.}}: TriDifNet: A Unified Three-Task Diffusion Network for End-to-End RGB-T Perception}

		\maketitle

		\begin{abstract}
			Infrared--visible (IR--VIS) multi-modal perception is important for robust scene understanding under challenging illumination conditions.
			In practice, infrared and visible images are often misaligned due to platform motion, viewpoint changes, and modality-dependent distortions, causing registration errors to propagate into subsequent fusion and detection stages.
			To address this issue, we propose TriDifNet, a unified three-task diffusion network for end-to-end RGB-T perception.
			To the best of our knowledge, TriDifNet is the first diffusion-based network that jointly optimizes registration, fusion, and detection for misaligned RGB-T pedestrian perception.
			
			TriDifNet consists of three diffusion-based modules, i.e., the Registration Diffusion Module (RegDM), the Fusion Diffusion Module (FusDM), and the Detection Diffusion Module (DetDM).
			RegDM aligns the unaligned infrared image to the optical reference by estimating a dense deformation field.
			FusDM performs residual conditional diffusion to generate a detection-oriented fused representation with optical structural cues and complementary infrared saliency.
			DetDM formulates pedestrian detection as an iterative denoising process in bounding-box space.
			To couple the three tasks, TriDifNet introduces a closed-loop optimization mechanism, where detection semantics guide fusion and registration, and the fusion result further regularizes geometric alignment.
			A stage-wise progressive training strategy and a lightweight Nash multi-task learning (Nash-MTL)-based late-stage refinement are further adopted to balance registration, fusion, and detection optimization.
			
			Extensive experiments on the KAIST multispectral pedestrian benchmark show that TriDifNet achieves superior downstream detection performance under misaligned RGB-T settings.
			Against a comprehensive set of strong system-level competitors,  TriDifNet achieves 65.44\% Recall, 57.55\% mAP$_{50}$, and 26.83\% mAP$_{50:95}$.

		\end{abstract}


		% \fi
		\IEEEpeerreviewmaketitle

		\section{Introduction}
		
		\IEEEPARstart{I}{nfrared--visible}(IR--VIS) multi-modal perception has become increasingly important in safety-critical scenarios such as night-time surveillance, autonomous driving, and search-and-rescue. Visible images provide rich texture and structural details, whereas infrared imagery highlights thermal saliency and remains robust under adverse illumination. In practice, however, the two modalities are often spatially misaligned due to platform motion, viewpoint variation, capture timing differences, and modality-dependent imaging distortions. The core challenge is therefore not merely the presence of cross-modal misalignment itself, but the fact that geometric alignment, cross-modal fusion, and task-level semantics are usually optimized in a decoupled manner. Recent studies have shown that registration and fusion should not be treated as entirely independent stages~\cite{p2_rfnet2022}, and that bidirectional interaction between them can improve both alignment quality and fusion fidelity~\cite{p2_rfdifnet2025}. More recent efforts have further explored tighter registration--fusion integration within unified or single-stage methods~\cite{p2_mulfscap2025,p2_imf2024}. For example, IMF introduces a one-stage progressive dense registration scheme and jointly optimizes it with a Transformer--Conv fusion subnetwork to mitigate structural distortions and edge ghosts in misaligned infrared--visible fusion~\cite{p2_imf2024}. Nevertheless, object detection in most existing registration--fusion systems still remains an external downstream consumer of the fused result rather than an endogenous optimization target inside the perception loop.
		
		A representative limitation appears in conventional sequential pipelines that perform registration first and fusion afterward. In such methods, registration is optimized before knowing which structures and regions are truly critical for downstream recognition, while fusion is forced to absorb the residual errors inherited from upstream geometric inaccuracies. This issue becomes more severe in practical misaligned IR--VIS scenarios, where explicit alignment is difficult~\cite{p3_defectreg2024} and cross-modal appearance discrepancy is large~\cite{p3_umfcmgr2022}. As a result, even mild misregistration can cause ghosting, double edges, structural distortion, or target deformation in the fused result~\cite{p3_reconet2022,p2_imf2024}. In complex real-world sensing conditions, such sensitivity to misalignment can further degrade the effectiveness of subsequent fusion-based perception~\cite{p3_earlyfire2025}. Consequently, the fusion stage is often reduced to a compensatory process on imperfectly aligned inputs, which limits both semantic fidelity and downstream robustness.
		
		Another line of work attempts to inject high-level semantics into fusion, for example by using segmentation or detection signals to guide the generation of fused representations~\cite{p4_seafusion2022,p4_detfusion2022,p4_metafusion2023,p4_psfusion2023}. Recent VIS--IR detection-oriented fusion methods further show that fusion and detection should be optimized in a task-consistent manner. For instance, MMI-Det explores multi-level visible--infrared integration with contour enhancement, fusion-focused feature interaction, contrastive bridging, and information-guided optimization to better balance fusion and detection objectives~\cite{p4_mmidet2024}. EI2Det further introduces cross-modal interaction, illumination-aware modality weighting, and edge-guided fusion to improve detection robustness under varying lighting conditions~\cite{p4_ei2det2025}. In parallel, recent weakly aligned RGB-T detection studies have recognized that strict cross-modal alignment is often unrealistic in practice~\cite{p5_cfdetr2024,p5_tsfadet2022,p5_lgcfa2022,p5_attentivealign2023}. Uncertainty-aware cross-modal detectors also indicate that practical RGB--IR perception suffers from object-level uncertainty caused by low-light visible observations, confusing infrared responses, thermal ghost shadows, and pose-induced cross-modal misalignment~\cite{p5_uacmdet2022}. Although these efforts demonstrate the value of task semantics, modality reliability, and alignment-aware detection, detection is still commonly treated either as an auxiliary guidance source for fusion or as a separate downstream task. As a result, it is rarely incorporated as a trainable component that jointly shapes alignment, fusion behavior, and final recognition. Consequently, what remains missing is a unified network in which geometric alignment, cross-modal fusion, and task-level detection semantics are optimized within a shared closed loop.
		
		Recently, diffusion models have demonstrated strong capacity for high-fidelity generation and restoration~\cite{LDM2022,SR3TPAMI2023,Palette2022,DDRM2022}, making them appealing for misaligned IR--VIS perception.
		More importantly, diffusion provides a unified iterative refinement view for the three tasks considered in this work.
		For registration, cross-modal matching under weak affine misalignment still involves spatially varying deformation estimation, which can be naturally cast as a progressive denoising process over deformation fields~\cite{DiffuseMorph2022,p2_rfdifnet2025}.
		For fusion, the optical image already provides a stable structural base, while the complementary infrared contribution is better modeled as a conditional residual restoration problem rather than direct full-image generation~\cite{DDFM2023,DDIF2024}.
		For detection, object localization can be reformulated as iterative denoising in the bounding-box space, enabling progressive refinement from noisy box hypotheses to final object predictions~\cite{DiffusionDet2023}.
		Therefore, instead of treating registration, fusion, and detection as heterogeneous modules coupled only at the loss level, diffusion offers a common optimization perspective in which geometry, cross-modal representation, and task semantics can all be refined through task-specific denoising processes.
		This shared formulation is particularly suitable for the proposed closed-loop design, as it makes the three tasks not only jointly trainable, but also conceptually unified under iterative error correction and mutual refinement.
		
		Building on this unified denoising perspective, \emph{we propose TriDifNet, a unified three-task diffusion network that jointly optimizes registration, fusion, and detection in an end-to-end manner}. Instead of following the conventional ``registration $\rightarrow$ fusion $\rightarrow$ detection'' chain, TriDifNet explicitly couples the three tasks within one optimization system. First, a \emph{Registration Diffusion Module} (RegDM) performs conditional diffusion to estimate a dense deformation field and warp the unaligned infrared image toward the optical reference. Then, a \emph{Fusion Diffusion Module} (FusDM) performs residual conditional diffusion under the joint guidance of the optical image and the registered infrared image. This residual parameterization is adopted since the optical image already provides a stable structural base, while the diffusion model only needs to generate the complementary cross-modal component that injects infrared saliency into target-relevant regions.
		
		To prevent detection from remaining a detached downstream head, we further elevate it into a trainable third task through a \emph{Detection Diffusion Module} (DetDM), which formulates object detection as an iterative denoising process in bounding-box space. DetDM operates on the detection-oriented fusion representation and produces the final pedestrian predictions, while simultaneously feeding task semantics back into the upstream modules. In particular, the fusion branch is constrained by an ROI-guided task-driven supervision, denoted as \emph{Det2Fus}, so that target-centered regions receive stronger infrared enhancement and more discriminative fusion behavior. In parallel, registration is regularized by infrared-domain object-level semantic consistency between the misaligned and registered infrared images, denoted as \emph{Det2Reg}. For this purpose, we employ a frozen YOLO detector only during training as a stable infrared-domain semantic anchor, rather than reusing the trainable fusion-side detector. This allows the registration branch to receive consistent object-level supervision without being affected by the optimization drift of the fusion-domain detection process. Finally, the fusion result is also fed back to regularize registration through a structural fusion-to-registration constraint, denoted as \emph{Fus2Reg}, forming a closed-loop optimization over the three tasks.
		To stabilize optimization across the three branches, TriDifNet
		is trained in a stage-wise progressive manner. After full closed-loop
		joint training, we further perform a lightweight Nash-MTL-based~\cite{nashmtl}
		late-stage refinement across the registration, fusion, and detection
		branches, so that the three task objectives can be dynamically balanced
		in the final optimization stage.
		
		In this way, TriDifNet is not a simple concatenation of three modules, but a jointly optimized perception network in which geometry, fusion, and semantics are mutually constrained. RegDM improves the geometric basis for both fusion and recognition; FusDM builds a cross-modal representation that is explicitly shaped by downstream task demands; and DetDM not only produces the final predictions but also acts as a semantic driver for upstream learning. Through this closed-loop design, the proposed network reduces early-stage error propagation and promotes coordinated improvement in alignment quality, cross-modal complementarity, and pedestrian detection performance.
		
		The main contributions of this work are summarized as follows:
		\begin{itemize}
			\item We propose a unified three-task end-to-end network for misaligned RGB-T pedestrian perception, which jointly optimizes infrared--visible registration, fusion, and detection within a shared closed loop instead of treating them as isolated or sequential stages.
			
			\item We develop a diffusion-based multi-module formulation for the three tasks, in which RegDM performs conditional deformation estimation for cross-modal alignment, FusDM performs residual conditional diffusion for complementary infrared--visible fusion, and DetDM formulates pedestrian detection as an iterative denoising process in bounding-box space.
			
			\item We design explicit cross-task coupling mechanisms, including \emph{Det2Reg} for detection-guided registration via a frozen infrared-domain semantic anchor, \emph{Det2Fus} for ROI-guided task-driven fusion, and \emph{Fus2Reg} for structural fusion-to-registration feedback, thereby establishing a genuine closed-loop optimization among geometry, fusion, and semantics.
			
			\item By integrating the above components into TriDifNet and employing a stage-wise training strategy with a lightweight late-stage three-branch Nash-MTL refinement, the proposed method achieves clear end-to-end detection improvements under misaligned RGB-T inputs.
			On the KAIST multispectral pedestrian benchmark with controlled RGB-T misalignment, TriDifNet reaches 65.44\% Recall, 57.55\% mAP$_{50}$, and 26.83\% mAP$_{50:95}$.
			The results show superior system-level performance against strong competing pipelines under the same setting.
		\end{itemize}
		
		% =========================================================
		% Preliminary -> Diffusion Models (aligned with Methodology, YOLO-head version)
		% (No heavy method details here; keep it as shared background + light instantiation)
		% =========================================================
		
		\section{Brief Introduction to Diffusion Model}
		\label{sec:preliminary}
		
		\label{subsec:prelim_diffusion}
		
		Diffusion models are likelihood-based generative models that learn a data distribution by simulating a gradual \emph{noising} process and then learning its \emph{denoising} reverse process~\cite{DDPM}.
		This section briefly introduces the standard DDPM formulation, including forward diffusion, reverse diffusion, and conditional diffusion.
		
		\paragraph{Forward diffusion}
		Let $\mathbf{x}_0 \sim q(\mathbf{x}_0)$ denote a clean target variable.
		A diffusion model defines a Markov forward process that gradually perturbs $\mathbf{x}_0$ into Gaussian noise:
		\begin{equation}
			q(\mathbf{x}_{1:T}\mid \mathbf{x}_0)=\prod_{t=1}^{T} q(\mathbf{x}_t \mid \mathbf{x}_{t-1})
			\label{eq:prelim_forward_chain}
		\end{equation}
		where each transition is commonly chosen as
		\begin{equation}
			q(\mathbf{x}_t \mid \mathbf{x}_{t-1})=\mathcal{N}\!\Big(\mathbf{x}_t;\sqrt{1-\beta_t}\,\mathbf{x}_{t-1},\,\beta_t \mathbf{I}\Big)
			\label{eq:prelim_forward_step}
		\end{equation}
		with $\{\beta_t\}_{t=1}^{T}$ a variance schedule.
		Define $\alpha_t=1-\beta_t$ and $\bar{\alpha}_t=\prod_{s=1}^{t}\alpha_s$.
		Then the marginal distribution admits a closed form:
		\begin{equation}
			\begin{aligned}
				q(\mathbf{x}_t\mid \mathbf{x}_0)
				&=\mathcal{N}\!\Big(\mathbf{x}_t;\sqrt{\bar{\alpha}_t}\,\mathbf{x}_0,\,(1-\bar{\alpha}_t)\mathbf{I}\Big), \\
				\mathbf{x}_t
				&=\sqrt{\bar{\alpha}_t}\,\mathbf{x}_0+\sqrt{1-\bar{\alpha}_t}\,\boldsymbol{\epsilon}
			\end{aligned}
			\label{eq:prelim_forward_marginal}
		\end{equation}
		where $\boldsymbol{\epsilon}\sim \mathcal{N}(\mathbf{0},\mathbf{I})$.
		
		\paragraph{Reverse diffusion}
		The generative reverse process learns to reconstruct $\mathbf{x}_0$ from $\mathbf{x}_T$:
		\begin{equation}
			p_\theta(\mathbf{x}_{0:T}\mid \mathbf{x}_T)=\prod_{t=1}^{T} p_\theta(\mathbf{x}_{t-1}\mid \mathbf{x}_t)
			\label{eq:prelim_reverse_chain}
		\end{equation}
		where each reverse step is parameterized as a Gaussian
		\begin{equation}
			p_\theta(\mathbf{x}_{t-1}\mid \mathbf{x}_t)=\mathcal{N}\!\Big(\mathbf{x}_{t-1};\boldsymbol{\mu}_\theta(\mathbf{x}_t,t),\,\boldsymbol{\sigma}^2_\theta(\mathbf{x}_t,t)\mathbf{I}\Big)
			\label{eq:prelim_reverse_step}
		\end{equation}
		Sampling starts from $\mathbf{x}_T \sim \mathcal{N}(\mathbf{0},\mathbf{I})$ and iteratively applies Eq.~\eqref{eq:prelim_reverse_step} to obtain $\mathbf{x}_0$.
		In practice, the denoiser is commonly trained to predict the injected noise $\boldsymbol{\epsilon}$, which leads to a simple and stable optimization objective.
		
		\paragraph{Conditional diffusion and guidance}
		For vision tasks, diffusion is typically used in a \emph{conditional} manner, where the reverse process is guided by an observation $\mathbf{y}$:
		\begin{equation}
			p_\theta(\mathbf{x}_{t-1}\mid \mathbf{x}_t,\mathbf{y})
			=\mathcal{N}\!\Big(\mathbf{x}_{t-1};\boldsymbol{\mu}_\theta(\mathbf{x}_t,t,\mathbf{y}),\,\boldsymbol{\sigma}^2_\theta(\mathbf{x}_t,t,\mathbf{y})\mathbf{I}\Big)
			\label{eq:prelim_cond_reverse}
		\end{equation}
		Compared with unconditional diffusion, conditional diffusion introduces task-related observations into the denoising process, enabling the model to generate or restore target variables under specific visual constraints.

		% If your paper uses citations in Preliminary, uncomment:
		% \cite{DualConditionDiffusionFusion, RFDifNet}

		\section{Methodology}
		\subsection{Overview}
		
\begin{figure*}[t]
			\centering
			\includegraphics[width=\linewidth]{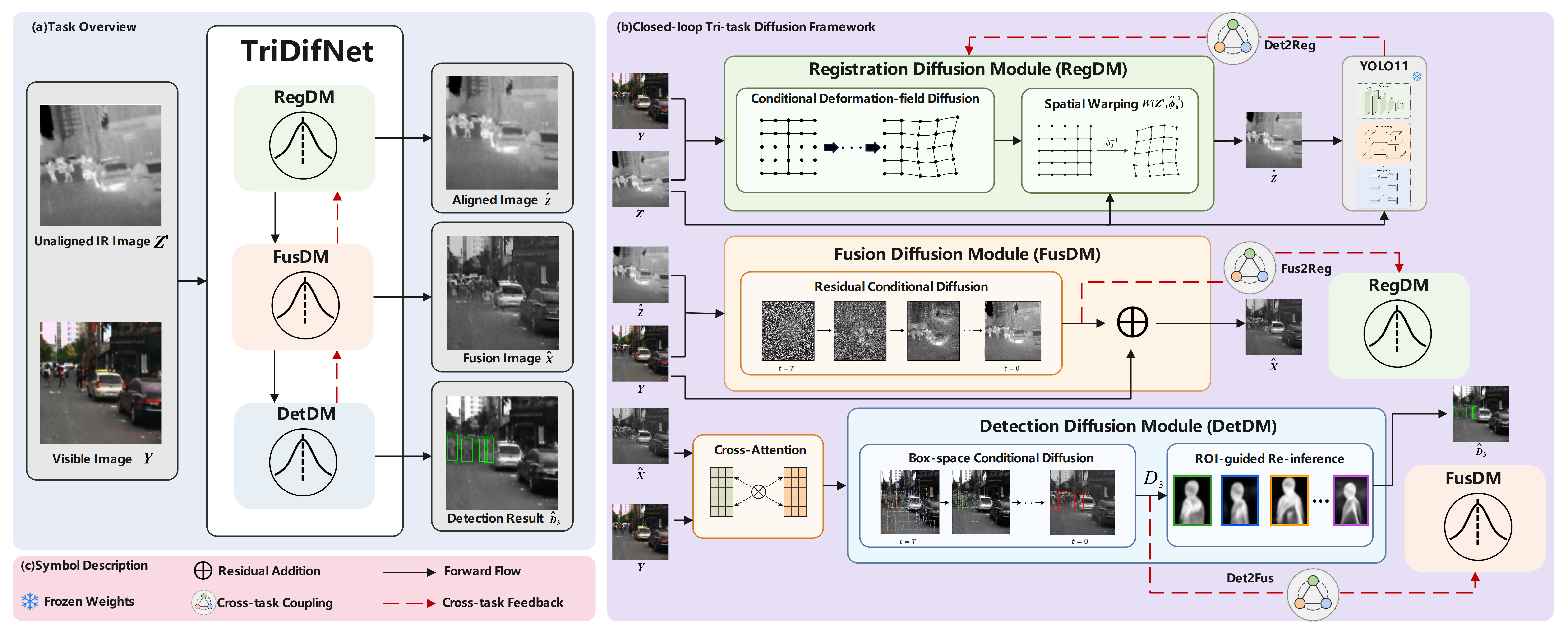}
			\caption{Overall architecture of TriDifNet.
				(a) Task overview. Given a visible image $Y$ and an unaligned infrared image $Z'$, TriDifNet sequentially produces the registered infrared image $\hat{Z}$, the fused image $\hat{X}$, and the final detection result $\hat{D}_{3}$.
				(b) Closed-loop tri-task diffusion network. RegDM estimates a dense deformation field through conditional deformation-field diffusion and spatial warping; FusDM generates a detection-oriented fused representation through residual conditional diffusion; and DetDM performs box-space conditional diffusion followed by ROI-guided re-inference.
				Black arrows denote the forward data flow, while red dashed arrows denote cross-task feedback paths, including Det2Reg, Fus2Reg, and Det2Fus.
				The frozen infrared-domain detector provides semantic guidance during training and is not updated with the TriDifNet parameters.
				(c) Symbol description.}
			\label{fig:framework}
			
		\end{figure*}

		This section describes the architecture and optimization of TriDifNet.
		As illustrated in Fig.~\ref{fig:framework}, TriDifNet is organized as a modular closed-loop network, where registration, fusion, and detection are executed sequentially in the forward path and are further coupled through cross-task feedback during training.
		
		TriDifNet consists of three core components:
		\begin{enumerate}
			\item \textbf{Registration Diffusion Module (RegDM):} estimates a dense deformation field to align the unaligned infrared image with the optical reference image, producing the registered infrared input for fusion.
			
			\item \textbf{Fusion Diffusion Module (FusDM):} performs residual conditional diffusion to generate a detection-oriented fused representation from the optical image and the registered infrared image.
			
			\item \textbf{Detection Diffusion Module (DetDM):} formulates object detection as a denoising diffusion process in bounding-box space and performs ROI-guided re-inference for final prediction.
		\end{enumerate}
		
		The three components are connected by explicit cross-task interactions.
		In the forward direction, RegDM provides aligned infrared guidance to FusDM (\emph{Reg-to-Fus}), and FusDM supplies the detector-oriented representation to DetDM (\emph{Fus-to-Det}).
		In the feedback direction, the fusion result regularizes registration through structural consistency (\emph{Fus2Reg}), a frozen infrared-domain detector provides semantic supervision for detection-guided registration (\emph{Det2Reg}), and the detector feeds ROI-guided task semantics back to the fusion branch (\emph{Det2Fus}).
		Through these interactions, the three tasks are jointly refined within a unified closed-loop network.
		
		The remainder of this section is organized as follows.
		Sec.~\ref{sec:notation_problem_formulation} defines the notation and formulates the three-task problem.
		Sec.~\ref{sec:regdm} details the Registration Diffusion Module (RegDM).
		Sec.~\ref{sec:fusdm} presents the Fusion Diffusion Module (FusDM).
		Sec.~\ref{sec:detdm} introduces the Detection Diffusion Module (DetDM).
		Sec.~\ref{sec:cross_task_coupling} describes the cross-task coupling paths among registration, fusion, and detection.
		Sec.~\ref{sec:training_strategy} presents the stage-wise training strategy and the Nash-MTL-based late-stage refinement.
		Finally, Sec.~\ref{sec:inference_pipeline} summarizes the inference pipeline.

		\subsection{Notation and Problem Formulation}
		\label{sec:notation_problem_formulation}
		
		Let $Y \in \mathbb{R}^{H \times W \times 3}$ denote the optical (visible) image, and let $Z' \in \mathbb{R}^{H \times W \times 1}$ denote the unaligned infrared image. 
		The goal of TriDifNet is to jointly optimize registration, fusion, and detection under the misaligned RGB-T setting.
		
		In the current work, the cross-modal discrepancy is modeled as controlled weak affine misalignment. Specifically, the infrared input is perturbed by in-plane rotation and translation with respect to the visible reference. This setting provides a practical weakly misaligned RGB-T scenario for evaluating whether registration, fusion, and detection can be optimized consistently within a unified network.
		
		Given $Y$ and $Z'$, the Registration Diffusion Module (RegDM) estimates an inverse deformation field
		\begin{equation}
			\hat{\phi}_0^{-1} = \mathrm{RegDM}(Y, Z')
		\end{equation}
		which is used to warp the unaligned infrared image into the optical reference coordinate system:
		\begin{equation}
			\hat{Z} = W(Z', \hat{\phi}_0^{-1})
		\end{equation}
		where $W(\cdot,\cdot)$ denotes the spatial warping operation. Here, $\hat{Z} \in \mathbb{R}^{H \times W \times 1}$ is the registered infrared image.
		
		Based on $Y$ and $\hat{Z}$, the Fusion Diffusion Module (FusDM) predicts a residual fusion component
		\begin{equation}
			\hat{X}_{\mathrm{res}} = \mathrm{FusDM}(Y, \hat{Z})
		\end{equation}
		and the final fused output is obtained by residual reconstruction
		\begin{equation}
			\hat{X} = Y + \hat{X}_{\mathrm{res}}
		\end{equation}
		where $\hat{X} \in \mathbb{R}^{H \times W \times C}$ denotes the fused representation.\footnote{The fused output contains both fusion-oriented and detection-oriented channels.}
		The detection-oriented fusion representation that is first fed into the Detection Diffusion Module (DetDM) is denoted by $F_{\mathrm{det}}$, which is obtained from the FusDM output after a visible--fusion cross-attention interaction module and serves as the first-pass detector input.
		
		The Detection Diffusion Module (DetDM) performs object detection in the bounding-box space. 
		Given $F_{\mathrm{det}}$, the detector first produces an initial prediction
		\begin{equation}
			D_3 = \mathrm{DetDM}(F_{\mathrm{det}})
		\end{equation}
		where $D_3$ denotes the first-pass detection result on the fusion-side detector input. 
		Using the ground-truth box prior and the prediction-derived ROI prior, a mixed ROI mask $M_{\mathrm{mix}}$ is constructed, and the detector-oriented representation is further refined into the task-driven detector representation $F_{\mathrm{det}}^{\mathrm{td}}$, which is used for the task-driven re-inference stage of DetDM.
		Therefore, the final fusion-side detection can be written as
		\begin{equation}
			\hat{D}_3 = \mathrm{DetDM}\!\left(F_{\mathrm{det}}^{\mathrm{td}}\right)
		\end{equation}
		
		In addition to fusion-side detection, TriDifNet also introduces infrared-domain detection for registration guidance. 
		Specifically, a frozen infrared detector is applied to the unregistered and registered infrared images, yielding
		\begin{equation}
			D_1 = \mathrm{Det}_{\mathrm{IR}}(Z'),
			\qquad
			D_2 = \mathrm{Det}_{\mathrm{IR}}(\hat{Z})
		\end{equation}
		where $D_1$ and $D_2$ provide stable semantic supervision for detection-guided registration.
		
		Overall, TriDifNet can be summarized as a three-task closed-loop system:
		\begin{equation}
			(Y, Z') \xrightarrow{\mathrm{RegDM}} \hat{Z}
			\xrightarrow{\mathrm{FusDM}} \hat{X}
			\xrightarrow{\mathrm{DetDM}} \hat{D}_3
		\end{equation}
		with three cross-task coupling paths:
		\begin{itemize}
			\item $\mathrm{Det2Reg}$: infrared-domain semantic consistency from $D_1$ and $D_2$ regularizes RegDM;
			\item $\mathrm{Det2Fus}$: detection semantics derived from $D_3$ and $M_{\mathrm{mix}}$ reshape the fusion-side detector representation from $F_{\mathrm{det}}$ to $F_{\mathrm{det}}^{\mathrm{td}}$ and feed back to FusDM;
			\item $\mathrm{Fus2Reg}$: the fused representation further regularizes registration through structural feedback.
		\end{itemize}
		
		Accordingly, the objective of this work is not to optimize registration, fusion, and detection independently, but to jointly learn them within a unified end-to-end network in which geometry, cross-modal fusion, and task-level semantics are mutually constrained.

		\subsection{Registration Diffusion Module (RegDM)}
		\label{sec:regdm}
		
		\subsubsection{Conditional Diffusion for Deformation Field Estimation}
		
		The Registration Diffusion Module (RegDM) formulates dense infrared-to-optical alignment as a conditional diffusion denoising process over the inverse deformation field. 
		Given the optical reference image $Y$ and the unaligned infrared image $Z'$, RegDM aims to estimate an inverse deformation field that maps the source infrared image into the optical reference coordinate system.
		
		At diffusion step $t$, the model takes a noisy deformation representation $\phi_t^{-1}$ together with a registration condition tensor $\mathrm{cond}_R$ and predicts the clean inverse deformation field. 
		Under the current IR--optical setting, the condition tensor is constructed by concatenating the optical reference and the unaligned infrared input along the channel dimension:
		\begin{equation}
			\mathrm{cond}_R = \mathrm{Concat}(Y, Z') \in \mathbb{R}^{H \times W \times 4}
		\end{equation}

		The RegDM denoiser $f_{\phi}(\cdot)$ predicts the injected Gaussian noise in the inverse deformation diffusion process as
		\begin{equation}
			\hat{\epsilon}_R = f_{\phi}(\phi_t^{-1}, \mathrm{cond}_R, t)
		\end{equation}
		
		According to the standard diffusion reconstruction form, the clean inverse deformation field can then be recovered as
		\begin{equation}
			\hat{\phi}_0^{-1}
			=
			\frac{1}{\sqrt{\bar{\alpha}_t}}
			\left(
			\phi_t^{-1}
			-
			\sqrt{1-\bar{\alpha}_t}\,\hat{\epsilon}_{R}
			\right)
		\end{equation}
		where $\bar{\alpha}_t$ denotes the cumulative noise coefficient defined in Sec.~\ref{sec:preliminary}.
		
		The predicted inverse deformation field is then applied to the unaligned infrared image through a spatial warping operation:
		\begin{equation}
			\hat{Z} = \mathcal{W}(Z', \hat{\phi}_0^{-1})
		\end{equation}
		where $\mathcal{W}(\cdot,\cdot)$ denotes the spatial warping operation. 
		The resulting $\hat{Z}$ is expected to be geometrically aligned with the optical reference image $Y$ and subsequently serves as the infrared condition for FusDM.
		
		In this way, RegDM provides the geometric foundation of the overall network: it first reduces cross-modal misalignment, then supplies the registered infrared image to the fusion branch, and meanwhile remains open to semantic and structural feedback from the downstream tasks through the closed-loop optimization.
		
		\subsubsection{Registration Objective}
		
		RegDM is optimized with a composite objective that balances diffusion reconstruction, deformation smoothness, and cross-modal alignment consistency:
		\begin{equation}
			L_R = L_R^{\mathrm{Diff}} + \alpha L_R^{\mathrm{smooth}} + \beta L_R^{\mathrm{align}}
		\end{equation}
		
		Here, $L_R^{\mathrm{Diff}}$ denotes the diffusion reconstruction loss for denoising the inverse deformation field. 
		Following the standard diffusion training formulation, the denoiser is trained to predict the injected Gaussian noise:
		\begin{equation}
			L_R^{\mathrm{Diff}} = \mathbb{E}_{\phi_0^{-1},\,\epsilon,\,t}
			\left[
			\left\|
			\epsilon - f_{\phi}(\phi_t^{-1}, \mathrm{cond}_R, t)
			\right\|_2^2
			\right]
		\end{equation}
		where $\epsilon \sim \mathcal{N}(0, I)$ and $\phi_t^{-1}$ is obtained by the forward diffusion process from the clean inverse deformation field.
		
		The smoothness term regularizes the spatial continuity of the predicted deformation field:
		\begin{equation}
			L_R^{\mathrm{smooth}} = \left\| \nabla_x \hat{\phi}_0^{-1} \right\|_1 + \left\| \nabla_y \hat{\phi}_0^{-1} \right\|_1
		\end{equation}
		where $\nabla_x$ and $\nabla_y$ denote horizontal and vertical finite-difference operators, respectively. 
		This term discourages unrealistic discontinuities and stabilizes the registration process.
		
		The alignment term enforces cross-modal consistency between the registered infrared image and the optical reference:
		\begin{equation}
			L_R^{\mathrm{align}} = 1 - \mathrm{NCC}(\hat{Z}, Y_{\mathrm{gray}})
		\end{equation}
		where $Y_{\mathrm{gray}}$ denotes the grayscale projection of the optical image used for cross-modal similarity measurement, and $\mathrm{NCC}(\cdot,\cdot)$ is the normalized cross-correlation operator. 
		This term encourages the warped infrared image to be geometrically consistent with the visible reference while remaining robust to cross-modal intensity differences.
		
		Therefore, the overall role of RegDM is twofold: it explicitly estimates the dense geometric correction needed for misaligned RGB-T inputs, and it provides the aligned infrared representation required by the subsequent fusion and detection processes.

		\subsection{Fusion Diffusion Module (FusDM)}
		\label{sec:fusdm}
		
		\subsubsection{Residual Conditional Diffusion for Infrared--Visible Fusion}
		
		After RegDM produces the registered infrared image $\hat{Z}$, the Fusion Diffusion Module (FusDM) performs conditional diffusion to generate a cross-modal residual representation under the joint guidance of the optical image and the aligned infrared image. 
		Unlike direct full-image generation, FusDM adopts a residual formulation since the optical image $Y$ already provides a stable structural and textural base, while the fusion branch mainly needs to inject complementary infrared saliency into target-relevant regions.
		
		Formally, the fusion condition tensor is defined as
		\begin{equation}
			\mathrm{cond}_F = \mathrm{Concat}(Y, \hat{Z})
		\end{equation}
		
		At diffusion step $t$, the FusDM denoiser $f_{\psi}(\cdot)$ takes the noisy residual fusion state $x_t$ together with $\mathrm{cond}_F$ and predicts the clean residual component:
		\begin{equation}
			\hat{X}_{\mathrm{res}} = f_{\psi}(x_t, \mathrm{cond}_F, t)
		\end{equation}
		
		The final fused representation is reconstructed by residual addition:
		\begin{equation}
			\hat{X} = Y + \hat{X}_{\mathrm{res}}
		\end{equation}
		where $\hat{X} \in \mathbb{R}^{H \times W \times C}$ denotes the fused output of FusDM. 
		This residual design explicitly preserves the dominant optical structures while allowing the diffusion model to focus on modality-complementary information from the infrared branch.
		
		Therefore, FusDM serves as the cross-modal representation hub of the overall network: it receives the geometrically corrected infrared signal from RegDM, produces the fused representation for image-level perception, and simultaneously supplies the detector-oriented fusion cue for the downstream detection branch.
		
		\subsubsection{Detector-Oriented Fusion Representation}
		
		In addition to generating the fused output $\hat{X}$, FusDM also provides a detector-oriented representation for DetDM. Rather than using a separate post-processing branch, this representation is built directly within the fusion pipeline, so that it can be refined by downstream detection semantics.
		
		Specifically, let $\tilde{F}_{\mathrm{det}} = \hat{X}^{(4)}$ denote the raw detection-oriented fusion channel extracted from the FusDM output, where $\hat{X}^{(4)}$ is the detector-side single-channel output before visible--fusion interaction.
		To preserve detector-friendly optical structural cues, we introduce a visible--fusion cross-attention module to generate the first-pass detector input $F_{\mathrm{det}}$. This helps keep the detector input closer to the original optical-image setting of DiffusionDet~\cite{DiffusionDet2023}, while complementing the subsequent ROI-guided infrared enhancement stage.
		
		We first encode the optical image and the raw detector-oriented fusion channel as
		\begin{equation}
			E_{\mathrm{vis}}=\phi_{\mathrm{vis}}(Y), 
			\qquad
			E_{\mathrm{fus}}=\phi_{\mathrm{fus}}(\mathrm{Rep3}(\tilde{F}_{\mathrm{det}}))
		\end{equation}
		where $\phi_{\mathrm{vis}}(\cdot)$ and $\phi_{\mathrm{fus}}(\cdot)$ denote the optical and fusion feature encoders, respectively, and $\mathrm{Rep3}(\cdot)$ denotes the channel-replication operator that expands a single-channel fusion map into a three-channel tensor for detector-side processing.
		
		Based on these two feature embeddings, the visible-guided cross-attention is formulated as
		\begin{equation}
			Q = W_Q E_{\mathrm{vis}}, 
			\qquad
			K = W_K E_{\mathrm{fus}}, 
			\qquad
			V = W_V E_{\mathrm{fus}}
		\end{equation}
		where $W_Q$, $W_K$, and $W_V$ are learnable linear projection operators for query, key, and value, respectively. 
		The attention map is computed by
		\begin{equation}
			A_{\mathrm{vf}} = \mathrm{Softmax}\!\left(\frac{QK^\top}{\sqrt{d}}\right)
		\end{equation}
		where $d$ is the feature dimension. 
		The detector-oriented fusion representation after visible--fusion interaction is then obtained as
		\begin{equation}
			F_{\mathrm{det}} = \mathcal{P}\!\left(E_{\mathrm{vis}} + A_{\mathrm{vf}}V\right)
		\end{equation}
		where $\mathcal{P}(\cdot)$ denotes the projection operator that maps the cross-attended feature representation back to the detector input representation space.
		
		The resulting $F_{\mathrm{det}}$ serves as the first-pass detector input for DetDM. 
		Therefore, the fusion branch provides not only the fused image representation $\hat{X}$ for image-level perception, but also a cross-attention-enhanced detector-oriented representation $F_{\mathrm{det}}$ that bridges fusion and detection.
		
		In this sense, FusDM plays two roles in TriDifNet:
		\begin{itemize}
			\item it produces the fused image representation $\hat{X}$ for cross-modal perception;
			\item it generates the detector-oriented representation $F_{\mathrm{det}}$ that bridges fusion and detection.
		\end{itemize}
		This design ensures that the fusion branch is not only optimized for visual fusion quality, but also prepared for semantic feedback from the downstream detector.
		
		\subsubsection{Fusion Objective}
		
		The base optimization of FusDM is defined as
		\begin{equation}
			L_F = L_F^{\mathrm{Diff}} + \lambda_1 L_F^{\mathrm{pix}} + \lambda_2 L_F^{\mathrm{str}}
		\end{equation}
		where $L_F^{\mathrm{Diff}}$ is the diffusion reconstruction loss for the residual fusion process, $L_F^{\mathrm{pix}}$ enforces pixel-level fidelity between the fused output and the fusion target, and $L_F^{\mathrm{str}}$ preserves structural consistency in the fused representation.
		
		The diffusion term is written as
		\begin{equation}
			L_F^{\mathrm{Diff}} =
			\mathbb{E}_{x_0,\epsilon,t}
			\left[
			\left\|
			\epsilon - f_{\psi}(x_t,\mathrm{cond}_F,t)
			\right\|_2^2
			\right]
		\end{equation}
		where $\epsilon \sim \mathcal{N}(0,I)$ and $x_t$ is the noisy residual state sampled from the forward diffusion process. 
		The pixel-level fidelity term is defined as
		\begin{equation}
			L_F^{\mathrm{pix}} = \left\| \hat{X} - X_{\mathrm{gt}} \right\|_1
		\end{equation}
		where $X_{\mathrm{gt}}$ denotes the fusion supervision target. 
		The structural regularization term is defined as
		\begin{equation}
			L_F^{\mathrm{str}} = 1 - \mathrm{SSIM}(\hat{X}, X_{\mathrm{gt}})
		\end{equation}
		which encourages the fused output to preserve perceptually meaningful scene structures.
		
		It should be noted that the above objective only defines the base fusion optimization of FusDM. In the full three-task network, this objective is further augmented by the task-driven Det2Fus constraints and the closed-loop cross-task feedback. The corresponding cross-task coupling mechanisms are formulated in Sec.~\ref{sec:cross_task_coupling}. With these additional couplings, the fusion branch is ultimately optimized not only for image-level fusion quality but also for detection-oriented semantic utility.

		\subsection{Detection Diffusion Module (DetDM)}
		\label{sec:detdm}
		
		\subsubsection{Diffusion-Based Detection Formulation}
		To prevent detection from remaining a detached downstream task, TriDifNet introduces a Detection Diffusion Module (DetDM) as the third trainable component in the closed-loop system. Instead of directly predicting object boxes in a one-shot manner, DetDM formulates pedestrian detection as an iterative denoising process in the bounding-box space. Given a detector-oriented fusion representation, the detector starts from a set of noisy box proposals and progressively refines them into the final object predictions.
		
		Given the first-pass detector input $F_{\mathrm{det}}$ obtained from the cross-attention-enhanced fusion representation, DetDM first generates an initial fusion-side prediction:
		\begin{equation}
			D_3 = \mathrm{DetDM}(F_{\mathrm{det}})
		\end{equation}
		where $D_3$ contains the first-pass pedestrian predictions, including the box locations and confidence scores on the fusion-side detector representation.
		
		\subsubsection{ROI-Guided Re-inference}
		\label{sec:roi_guided_reinference}
		
		Unlike a conventional single-pass detector, the current DetDM does not directly use the first-pass prediction $D_3$ as the final output. 
		Instead, $D_3$ is converted into an ROI prior and used to refine the detector-side fusion representation for second-pass detection.
		
		\begin{figure}[!t]
			\centering
			\includegraphics[width=0.92\columnwidth]{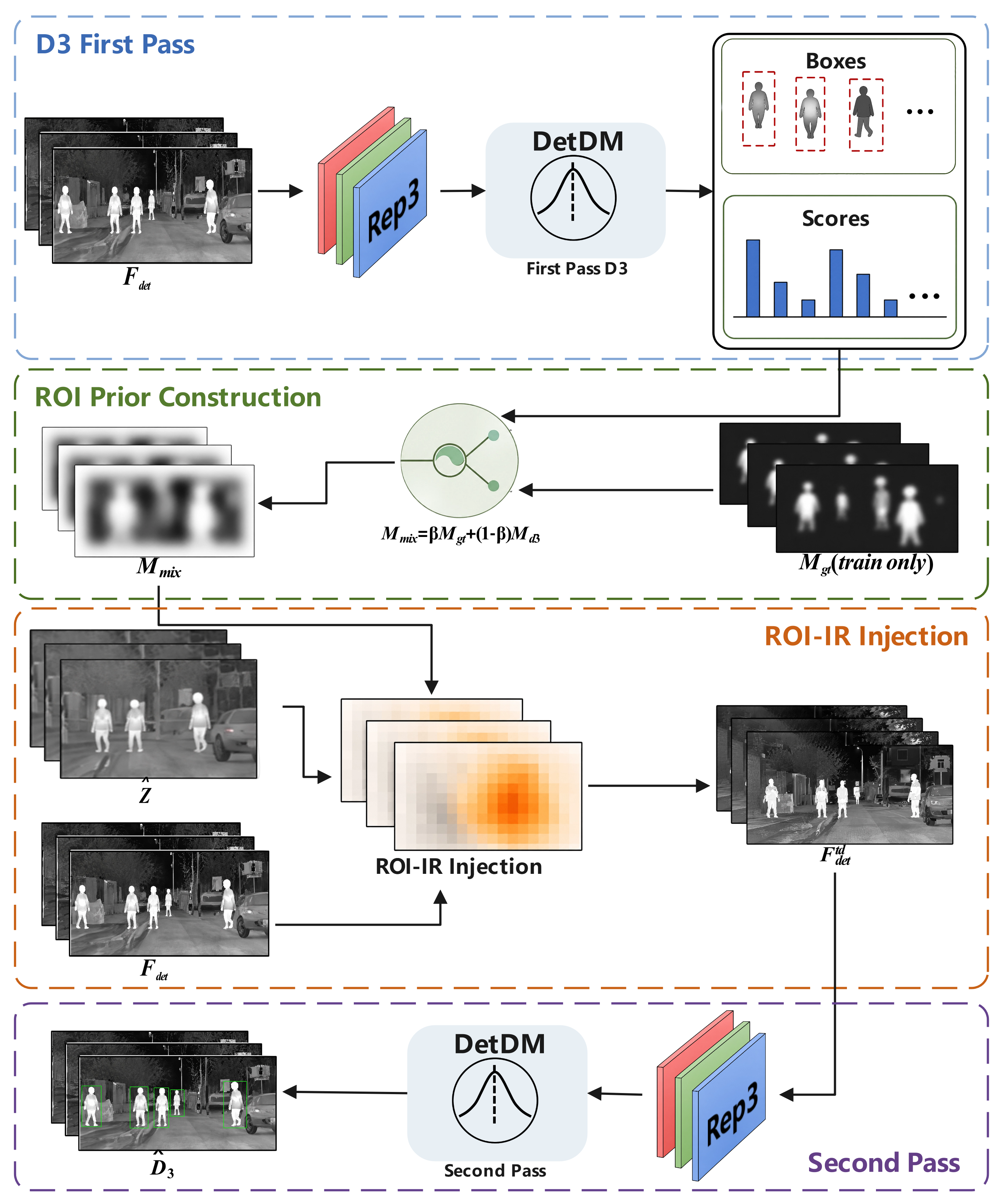}
			\caption{ROI-guided re-inference in DetDM. 
				First-pass predictions are converted into an ROI prior, which guides ROI-IR injection for second-pass detection.}
			\label{fig:roi_reinference}
		\end{figure}

		As shown in Fig.~\ref{fig:roi_reinference}, this re-inference process consists of four steps: D3 first-pass detection, ROI prior construction, ROI-IR injection, and second-pass detection. 
		To construct the task-driven refinement prior, the ground-truth boxes are first converted into a soft ROI mask:
		\begin{equation}
			M_{\mathrm{gt}} = \mathcal{R}(B_{\mathrm{gt}})
		\end{equation}
		where $B_{gt}$ denotes the ground-truth bounding-box annotations, and $R(\cdot)$ denotes the ROI mask generation operator that converts box annotations into a soft ROI mask.
		Similarly, the first-pass prediction $D_3$ is converted into a prediction-derived ROI mask:
		\begin{equation}
			M_{d3} = \mathcal{R}(D_3)
		\end{equation}
		
		To balance supervision from the ground-truth prior and the detector prediction prior, a mixed ROI mask is constructed as
		\begin{equation}
			M_{\mathrm{mix}} = \beta M_{\mathrm{gt}} + (1-\beta)M_{d3}
		\end{equation}
		where $\beta$ is an adaptive mixing coefficient determined by the agreement between the prediction-derived ROI and the ground-truth ROI. In practice, the agreement is measured by the mask-level IoU between $M_{d3}$ and $M_{\mathrm{gt}}$, and $\beta$ is constrained within a GT-dominant interval to prevent the mixed ROI prior from being overly dominated by unstable early predictions.
		
		The mixed ROI mask is then used to refine the first-pass detector input $F_{\mathrm{det}}$ into a task-driven detector representation:
		\begin{equation}
			F_{\mathrm{det}}^{\mathrm{td}} = \mathcal{T}(F_{\mathrm{det}}, \hat{Z}, M_{\mathrm{mix}})
		\end{equation}
		where $\mathcal{T}(\cdot)$ denotes the ROI-guided task-driven rewriting operator. 
		Intuitively, this operation preserves the original detector-oriented fusion response in non-target regions while injecting stronger infrared saliency into target-centered ROI regions. 
		As a result, the detector input is no longer a purely fusion-driven representation, but a task-aware representation explicitly reshaped by the object prior.
		
		The refined representation is then used for the second-pass detection:
		\begin{equation}
			\hat{D}_3 = \mathrm{DetDM}(F_{\mathrm{det}}^{\mathrm{td}})
		\end{equation}
		where $\hat{D}_3$ denotes the final fusion-side detection output.
		
		\subsubsection{Detection Objective}
		
		The optimization of DetDM consists of two coupled levels. 
		At the detector level, we adopt the box-space diffusion detection objective to supervise the final second-pass prediction $\hat{D}_3$:
		\begin{equation}
			L_D = L_D^{\mathrm{cls}} + L_D^{\mathrm{box}} + L_D^{\mathrm{giou}}
		\end{equation}
		where $L_D^{\mathrm{cls}}$ is the classification loss, $L_D^{\mathrm{box}}$ is the box regression loss, and $L_D^{\mathrm{giou}}$ is the generalized intersection-over-union loss.
		
		At the representation level, the detector is further coupled with the fusion branch through ROI-guided task-driven constraints. 
		These constraints do not merely supervise the final boxes; instead, they act on the detector-oriented fusion representation itself through $M_{\mathrm{mix}}$ and $F_{\mathrm{det}}^{\mathrm{td}}$. 
		Consequently, the detector not only predicts pedestrian instances, but also provides semantic feedback that reshapes the fusion-side representation before final inference.
		
		Therefore, the overall role of DetDM in TriDifNet is twofold:
		\begin{itemize}
			\item it serves as the final pedestrian detector operating on fusion-side representations;
			\item it acts as a task-semantic driver that constructs the ROI-guided feedback used by the Det2Fus pathway.
		\end{itemize}
		
		In this way, detection is lifted from a passive downstream consumer to an endogenous component of the closed-loop optimization, enabling the final detector to jointly influence fusion behavior and, through the cross-task constraints, the overall RGB-T perception process.

		\subsection{Cross-Task Coupling: Registration--Fusion--Detection}
		\label{sec:cross_task_coupling}
		
		A key contribution of TriDifNet is to explicitly couple registration, fusion, and detection in a closed-loop manner, so that improvements in one task can directly constrain and enhance the others. 
		The resulting interactions include a forward coupling from registration to fusion, a semantic feedback path from detection to registration, a task-driven feedback path from detection to fusion, and a structural feedback path from fusion to registration. 
		These couplings are denoted as \emph{Reg-to-Fus}, \emph{Fus2Reg}, \emph{Det2Reg},  \emph{Det2Fus}, and \emph{Fus-to-Det}, respectively.
		
		\subsubsection{Registration-to-Fusion Consistency (Reg-to-Fus)}
		
		The registered infrared image $\hat{Z}$ produced by RegDM serves as the geometrically aligned infrared condition for FusDM. 
		By replacing the raw unaligned infrared input with $\hat{Z}$, FusDM operates on a cross-modal pair with substantially reduced spatial discrepancy, which stabilizes the residual diffusion process and improves the structural consistency of the fused output. 
		Accordingly, Reg-to-Fus is implemented by using $\hat{Z}$ as the infrared condition in the FusDM residual diffusion process defined in Sec.~\ref{sec:fusdm}.
		This forward coupling allows FusDM to operate on a geometrically corrected RGB-T pair and stabilizes the generation of the fused representation.
		
		\subsubsection{Fusion-to-Registration Structural Feedback (Fus2Reg)}
		
		Beyond serving as the output of FusDM, the fused representation also feeds back to RegDM as a structural regularizer. 
		The motivation is that the fusion result encodes cross-modal structural consistency after complementary information from both modalities has been integrated, and such fused structural cues can be used to further regularize geometric alignment.
		
		We denote this feedback path as \emph{Fus2Reg}. 
		To preserve geometry-sensitive structures, we implement Fus2Reg through a fractional-order derivative consistency between the registered infrared image and the structure extracted from the fused representation:
		\begin{equation}
			L_{\mathrm{Fus2Reg}}
			=
			\left\|
			D^{\alpha}(\hat{Z}) - D^{\alpha}\!\big(\mathrm{CA}(\hat{X})\big)
			\right\|_1
		\end{equation}
		where $D^{\alpha}(\cdot)$ denotes a fractional-order derivative operator with order $\alpha \in (0,1]$, and $\mathrm{CA}(\cdot)$ denotes channel-wise averaging that maps the fused representation to a comparable structural map. 
		This constraint encourages the registered infrared image to share consistent structural responses with the fusion-enhanced structure, thereby correcting deformation estimation using fusion-driven geometry cues.
		
		\subsubsection{Detection-Guided Registration (Det2Reg)}
		
		Low-level alignment cues alone are insufficient for object-centric RGB-T perception, since registration should also preserve the semantic consistency of target regions across the unaligned and registered infrared domains. 
		To this end, we introduce \emph{Det2Reg}, which uses a frozen infrared-domain detector as a stable semantic anchor during training.
		
		Specifically, the frozen detector is applied to the unaligned infrared image $Z'$ and the registered infrared image $\hat{Z}$, producing
		\begin{equation}
			D_1 = \mathrm{Det}_{\mathrm{IR}}(Z'),
			\qquad
			D_2 = \mathrm{Det}_{\mathrm{IR}}(\hat{Z})
		\end{equation}
		These two predictions provide object-level semantic supervision for registration, encouraging RegDM to align object-centric structures more accurately. 
		Unlike using the trainable fusion-side detector, this frozen infrared-domain detector provides a stable semantic reference and avoids introducing optimization drift from the fusion-side detection branch into the registration supervision.
		
		Accordingly, the Det2Reg constraint can be summarized as
		\begin{equation}
			L_{\mathrm{Det2Reg}} = L_{\mathrm{sem}}(D_1, D_2; B_{\mathrm{gt}})
		\end{equation}
		where $L_{\mathrm{sem}}(\cdot)$ denotes the infrared-domain semantic consistency objective induced by the frozen detector predictions and the ground-truth bounding-box annotations $B_{\mathrm{gt}}$. Specifically, we formulate it as
		\begin{equation}
			\begin{split}
				L_{\mathrm{sem}}(D_1, D_2; B_{\mathrm{gt}})
				=
				L_{\mathrm{det}}(D_2, B_{\mathrm{gt}}) \\
				+
				\lambda_{\mathrm{sc}}L_{\mathrm{sc}}(D_1, D_2; B_{\mathrm{gt}})
			\end{split}
		\end{equation}
		where $L_{\mathrm{det}}(D_2, B_{\mathrm{gt}})$ supervises the registered-infrared detection prediction $D_2$ with the ground-truth boxes, and $L_{\mathrm{sc}}(D_1, D_2; B_{\mathrm{gt}})$ encourages object-level semantic responses before and after registration to remain consistent around annotated target regions. In TriDifNet, this term acts as a training-time semantic regularizer for RegDM and is not used during inference.
		
		In practice, the semantic consistency term is computed on detector responses matched to the ground-truth boxes:
		\begin{equation}
			L_{\mathrm{sc}}(D_1,D_2;B_{\mathrm{gt}})
			=
			\frac{1}{|B_{\mathrm{gt}}|}
			\sum_{j=1}^{|B_{\mathrm{gt}}|}
			\left\|
			\mathbf{s}_{2}^{j}
			-
			\operatorname{sg}(\mathbf{s}_{1}^{j})
			\right\|_1
		\end{equation}
		where $\mathbf{s}_{1}^{j}$ and $\mathbf{s}_{2}^{j}$ denote the detector semantic responses associated with the $j$-th ground-truth box from $D_1$ and $D_2$, respectively, and $\operatorname{sg}(\cdot)$ denotes the stop-gradient operation.
		
		\subsubsection{Detection-Aware Fusion (Det2Fus)}
		
		To explicitly inject object-level semantics into the fusion process, we further design \emph{Det2Fus}, which allows detection to reshape the detector-oriented fusion representation before final inference. 
		Unlike simply attaching a detection loss to the fused output, Det2Fus first constructs an ROI-guided prior and uses it to rewrite the detector-side fusion representation. 
		The construction of the ROI masks $M_{\mathrm{gt}}$, $M_{\mathrm{d3}}$, the mixed prior $M_{\mathrm{mix}}$, and the task-driven detector representation $F_{\mathrm{det}}^{\mathrm{td}}$ has been introduced in Sec.~\ref{sec:roi_guided_reinference}.
		Here, we focus on how this detector-side semantic prior is further fed back to FusDM through task-driven fusion constraints.
		
		\begin{figure}[!b]
			\centering
			\includegraphics[width=0.98\columnwidth]{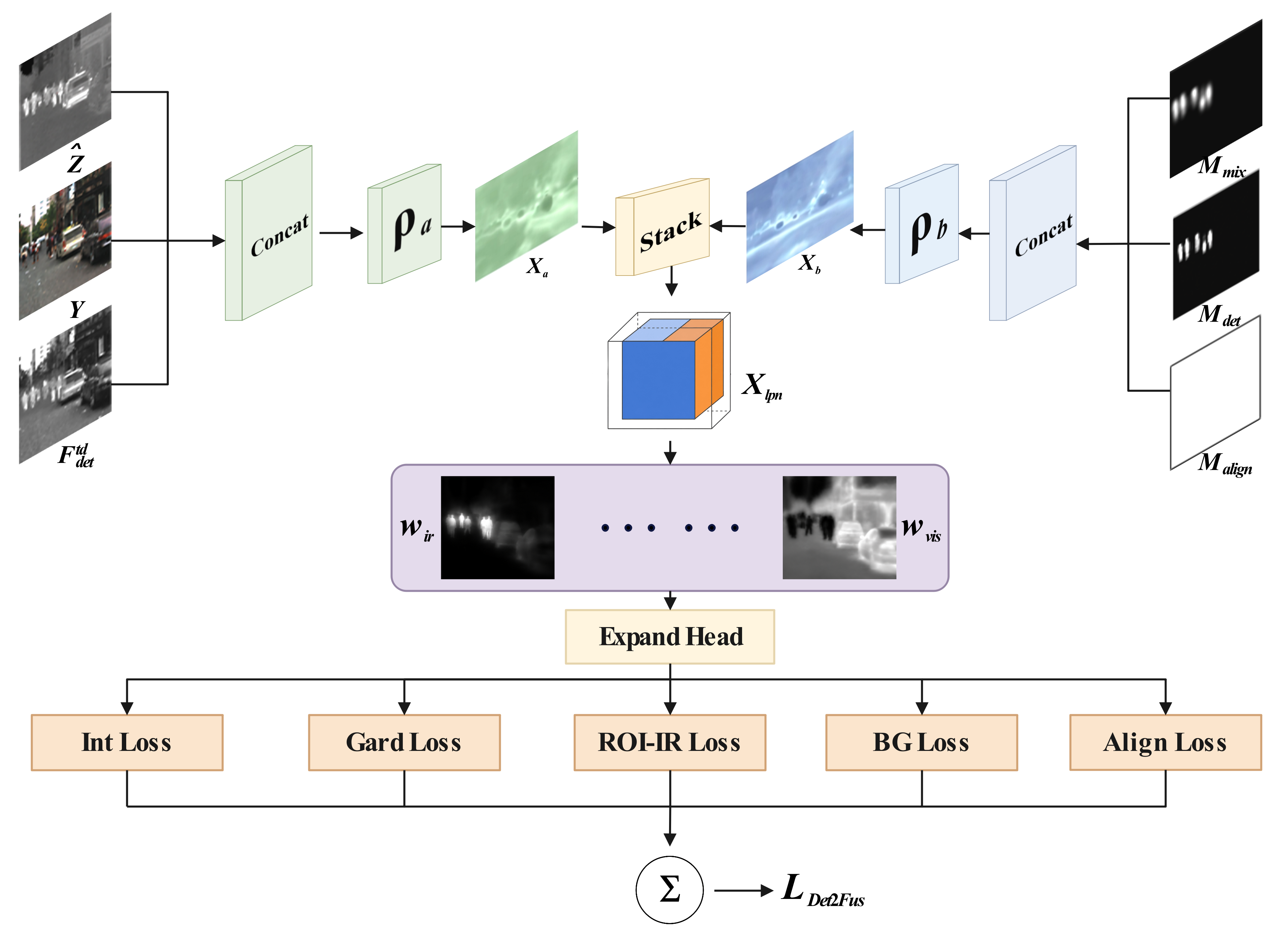}
			\caption{TDFusion-style loss-side guidance branch in Det2Fus. 
				It predicts task-driven spatial weights for ROI-aware fusion constraints during training.}
			\label{fig:det2fus_tdfusion}
		\end{figure}
		
		As shown in Fig.~\ref{fig:det2fus_tdfusion}, we introduce a TDFusion-style learnable weighting branch to strengthen the detector-to-fusion feedback. 
		This branch serves as a training-time loss-side guidance module rather than an additional inference-time fusion generator. 
		Specifically, a learnable weighting generator $\mathcal{G}_{\mathrm{td}}$ takes the registered infrared image, the optical image, the first-pass detector input, and the mixed ROI prior as inputs:
		\begin{equation}
			\begin{split}
				\{w_{\mathrm{ir}}, w_{\mathrm{vis}}, w_{\mathrm{roi}}, w_{\mathrm{edge}}, 
				w_{\mathrm{bg}}, w_{\mathrm{align}}\}
				= \\
				\mathcal{G}_{\mathrm{td}}(\hat{Z}, Y, F_{\mathrm{det}}^{\mathrm{td}}, M_{\mathrm{mix}})
			\end{split}
		\end{equation}
		where the predicted weights are modality-aware and region-aware spatial maps that adaptively control infrared enhancement, visible suppression, ROI emphasis, edge preservation, background inhibition, and alignment relaxation.
		
		To avoid a trivial visible-dominant solution, the TDFusion branch biases the effective weighting toward infrared responses inside ROI regions while suppressing unnecessary visible dominance in object-centric areas. 
		Therefore, it regularizes FusDM through the detector-aware representation and the ROI-guided semantic prior without introducing an additional fused-image generation path during inference.
		
		With the ROI-guided rewriting and the TDFusion-style weighting branch, the Det2Fus objective imposes task-driven constraints directly on the detector-oriented fusion representation:
		\begin{equation}
			\begin{split}
				L_{\mathrm{Det2Fus}} =\;&	L_{\mathrm{int}} +
				\lambda_g\,L_{\mathrm{grad}} + \lambda_r\,L_{\mathrm{roi\mbox{-}ir}} \\
				&+ \lambda_b\,L_{\mathrm{bg}} + \lambda_a\,L_{\mathrm{align}}
			\end{split}
		\end{equation}
		where all terms are computed on the ROI-rewritten detector-side fusion representation $F_{\mathrm{det}}^{\mathrm{td}}$.
		$L_{\mathrm{int}}$ enforces weighted intensity consistency with the registered infrared image $\hat{Z}$ and the visible structural proxy $\mathrm{CA}(Y)$.
		$L_{\mathrm{grad}}$ preserves the dominant Sobel-gradient responses from the two modalities.
		$L_{\mathrm{roi\mbox{-}ir}}$ applies an ROI-masked infrared consistency constraint to enhance target saliency.
		$L_{\mathrm{bg}}$ suppresses unnecessary detector-channel residual activation in background regions.
		$L_{\mathrm{align}}$ maintains consistency with the teacher detector-oriented fusion target while relaxing this constraint inside ROI regions.
		
		This design enables DetDM to shape the fusion behavior at the representation level while preserving global fusion fidelity outside object-centric regions. 
		Therefore, Det2Fus is not a simple auxiliary detection loss, but a representation-level semantic feedback mechanism that reshapes the detector-oriented fusion branch before final detection.
		
		\subsubsection{Fusion-Enhanced Detection(Fus-to-Det)}
		
		The fused representation also enhances detection in the forward direction. 
		Specifically, FusDM provides the first-pass detector input $F_{\mathrm{det}}$, while the refined representation $F_{\mathrm{det}}^{\mathrm{td}}$ is used for the final detector inference.
		In this way, the fusion branch not only receives semantic feedback from DetDM, but also produces the task-specialized detector input that drives the final diffusion-based detection process.
		
		\subsubsection{Overall Training Objective}
		
		The overall optimization of TriDifNet is jointly driven by registration, fusion, and detection objectives:
		\begin{equation}
			L_{\mathrm{total}} = L_R + L_F + L_D
		\end{equation}
		
		The registration objective is defined as
		\begin{equation}
			\begin{split}
				L_R
				=
				L_R^{\mathrm{Diff}}
				+\alpha L_R^{\mathrm{smooth}}
				+\beta L_R^{\mathrm{align}} \\
				+\lambda_{\mathrm{det2reg}} L_{\mathrm{Det2Reg}}
				+\lambda_{\mathrm{fus2reg}} L_{\mathrm{Fus2Reg}}
			\end{split}
		\end{equation}
		where the first three terms are the base registration losses of RegDM, and the last two terms represent semantic feedback from Det2Reg and structural feedback from Fus2Reg.

		The fusion objective is defined as
		\begin{equation}
			L_F = L_F^{\mathrm{Diff}} + \lambda_1 L_F^{\mathrm{pix}} + \lambda_2 L_F^{\mathrm{str}} + \lambda_{\mathrm{det2fus}} L_{\mathrm{Det2Fus}}
		\end{equation}
		where the first three terms denote the base fusion optimization of FusDM, and the last term introduces the ROI-guided task-driven semantic feedback from Det2Fus.
		
		In the current implementation, $L_{\mathrm{Det2Fus}}$ already includes the ROI-guided infrared enhancement and the TDFusion-style learnable weighting mechanism. 
		The corresponding TDFusion branch acts as a training-time loss-side guidance module, which shapes the optimization trajectory of FusDM without introducing an additional fused-image generation path at inference time.
		
		The detection objective $L_D$ is imposed on the final detector output driven by the refined detector representation $F_{\mathrm{det}}^{\mathrm{td}}$.
		It follows the box-space diffusion detection loss defined in Sec.~\ref{sec:detdm}, including classification, box regression, and generalized intersection-over-union terms.
		
		Through this formulation, RegDM maintains geometric consistency, FusDM is jointly shaped by image-level supervision and detection-aware semantic feedback, and DetDM is optimized on the detector-oriented fusion representation refined by ROI-guided re-inference. 
		As a result, the final training strategy consistently reflects the current three-task closed-loop network: geometry, fusion, and semantics are no longer optimized in isolation, but are mutually constrained within one unified end-to-end network.
		The above formulation defines the branch-wise optimization
		targets of TriDifNet. The practical training
		schedule used to optimize these objectives, including the
		late-stage Nash-MTL-based refinement, is introduced next.
		
		\subsection{Training Strategy and Nash-MTL-based Late-Stage Refinement}
		\label{sec:training_strategy}
		
		TriDifNet is not trained with a single static
		schedule from the beginning. Instead, we adopt a stage-wise
		progressive training strategy to stabilize optimization across
		registration, fusion, and detection. Specifically, the training
		process consists of multiple stages. We first optimize the
		registration branch while freezing the fusion and detection
		branches, so that RegDM can establish a reliable geometric
		alignment basis. Next, we optimize the coupled RegDM--FusDM
		and RegDM--DetDM settings separately by freezing one branch
		at a time, which allows the fusion and detection branches to
		adapt to the registration-guided inputs under reduced
		optimization interference. After that, all three branches are
		jointly optimized in the full closed-loop setting, so that
		Det2Reg, Det2Fus, and Fus2Reg can simultaneously
		participate in the end-to-end interaction.
		
		After the above joint training stage, we further introduce
		a late-stage refinement step to improve the optimization
		balance among the three task objectives. Unlike a fixed
		loss summation, Nash multi-task learning (Nash-MTL) is used
		as a dynamic task-coordination strategy to adaptively balance
		registration, fusion, and detection in the final training stage.
		Let
		\begin{equation}
			\mathcal{L}_{\mathrm{act}} = \{L_R, L_F, L_D\}
		\end{equation}
		denote the set of active trainable objectives in this refinement stage.
		We compute their gradients with respect to a shared intermediate representation $\mathbf{s}$
		and form the gradient matrix
		\begin{equation}
			\mathbf{G}=
			\left[
			\nabla_{\mathbf{s}}L_R,\,
			\nabla_{\mathbf{s}}L_F,\,
			\nabla_{\mathbf{s}}L_D
			\right]^{\top}
		\end{equation}
		Following Nash-MTL, the dynamic task weights
		$\boldsymbol{\alpha}=[\alpha_R,\alpha_F,\alpha_D]^{\top}$ are obtained by solving
		\begin{equation}
			\mathbf{G}\mathbf{G}^{\top}\boldsymbol{\alpha}
			=
			\mathbf{1}/\boldsymbol{\alpha}
		\end{equation}
		where $\mathbf{1}/\boldsymbol{\alpha}$ denotes the element-wise
		reciprocal. The resulting weighted objective for the late-stage
		refinement is
		\begin{equation}
			L_{\mathrm{Nash}}=\alpha_R L_R+\alpha_F L_F+\alpha_D L_D
		\end{equation}
		
		This design is motivated by the observation that the final
		closed-loop optimization contains heterogeneous objectives
		from geometric alignment, cross-modal fusion, and task-level
		detection. Directly summing these losses with fixed weights
		may cause one task to dominate the optimization and suppress
		the others. Nash-MTL provides a dynamic balancing mechanism
		that coordinates the gradients of the three branches and
		encourages a more stable multi-task update. If a training batch
		does not contain valid detection supervision, the detector
		branch is excluded from the bargaining step, and the
		optimization falls back to the available active objective(s)
		rather than enforcing an invalid three-task update.
		
		It should be emphasized that Nash-MTL is used only in this
		training-time late-stage refinement step. It improves the
		optimization balance among RegDM, FusDM, and DetDM, but
		does not introduce additional inference-time modules or change
		the structural closed-loop coupling of TriDifNet.

		\subsection{Inference Pipeline}
		\label{sec:inference_pipeline}
		
		During inference, TriDifNet performs end-to-end RGB-T perception through a sequential forward pass over RegDM, FusDM, and DetDM, while preserving the detector-side ROI-guided re-inference mechanism. 
		Given an optical image $Y$ and an unaligned infrared image $Z'$, RegDM first estimates the inverse deformation field and warps the infrared image into the optical reference coordinate system:
		\begin{equation}
			\hat{\phi}_0^{-1} = \mathrm{RegDM}(Y, Z'),
			\qquad
			\hat{Z} = \mathcal{W}(Z', \hat{\phi}_0^{-1})
		\end{equation}
		The registered infrared image $\hat{Z}$ is then fed into FusDM together with the optical image $Y$ to produce the fused representation:
		\begin{equation}
			\hat{X}_{\mathrm{res}} = \mathrm{FusDM}(Y, \hat{Z}),
			\qquad
			\hat{X} = Y + \hat{X}_{\mathrm{res}}
		\end{equation}

		Next, the visible--fusion cross-attention module converts the raw detector-side fusion channel into the cross-attention-enhanced detector-oriented representation $F_{\mathrm{det}}$, which serves as the first-pass detector input.
		\begin{equation}
			F_{\mathrm{det}} = \mathcal{P}\!\left(E_{\mathrm{vis}} + A_{\mathrm{vf}}V\right)
		\end{equation}
		
		Based on $F_{\mathrm{det}}$, DetDM performs an initial diffusion-based detection and produces the first-pass prediction
		\begin{equation}
			D_3 = \mathrm{DetDM}(F_{\mathrm{det}})
		\end{equation}
		The first-pass prediction is then converted into a prediction-derived ROI prior
		\begin{equation}
			M_{d3} = \mathcal{R}(D_3)
		\end{equation}
		where $\mathcal{R}(\cdot)$ denotes the ROI mask generation operator from the detector prediction. 
		Using this prediction-derived ROI prior, the detector-oriented representation is rewritten into a task-driven refined detector input:
		\begin{equation}
			F_{\mathrm{det}}^{\mathrm{td}} = \mathcal{T}(F_{\mathrm{det}}, \hat{Z}, M_{d3})
		\end{equation}
		where $\mathcal{T}(\cdot)$ denotes the ROI-guided rewriting operator that enhances infrared saliency in target-centered regions while preserving the original fusion-driven detector response in non-target regions. 
		The final pedestrian detection is then obtained through ROI-guided re-inference:
		\begin{equation}
			\hat{D}_3 = \mathrm{DetDM}(F_{\mathrm{det}}^{\mathrm{td}})
		\end{equation}
		
		It is worth noting that the training-only semantic and structural supervision terms are not used during inference. 
		In particular, the frozen infrared-domain YOLO detector employed by Det2Reg serves only as a semantic anchor during training and is removed at test time. 
		Similarly, the loss terms associated with Det2Reg, Det2Fus, and Fus2Reg are used only for optimization and are not evaluated during inference. 
		However, the detector-side ROI-guided refinement remains part of the forward inference pipeline, since it directly constructs the task-driven detector representation $F_{\mathrm{det}}^{\mathrm{td}}$ from the first-pass prediction prior $M_{d3}$ for the second-pass detection.
		
		Therefore, the inference stage of TriDifNet outputs three perceptually and semantically meaningful results: the registered infrared image $\hat{Z}$, the fused representation $\hat{X}$, and the final pedestrian detection result $\hat{D}_3$. 
		This inference pipeline consistently reflects the design principle of TriDifNet: geometry is first corrected by RegDM, cross-modal information is integrated by FusDM, and final task-level semantics are produced by DetDM through prediction-guided detection refinement.
		
		\section{Experiments}
		
		\begin{table*}[t]
			\centering
			\caption{Comparison Settings For Different Evaluation Settings}
			\label{tab:comparison_protocols}
			\begin{tabular}{cccc}
				\Xhline{1.2pt}
				Evaluation Setting & Input & Output & Primary Goal \\
				\hline
				Registration comparison & Misaligned RGB-T & Registered IR & Image-level registration accuracy \\
				Fusion comparison & Aligned RGB-T & Fusion result & Image-level fusion effectiveness \\
				End-to-end comparison & Misaligned RGB-T & Final detection & Full-system perception performance \\
				\Xhline{1.2pt}
			\end{tabular}
		\end{table*}
		
		\subsection{Datasets, Misalignment Setting, and Evaluation Metrics}
		
		\paragraph{Datasets}
		All experiments are conducted on the KAIST Multispectral Pedestrian Benchmark~\cite{KAIST}, which provides paired visible--thermal images for RGB--T pedestrian perception.
		Following a cleaned annotation setting, heavily occluded, severely truncated, extremely small, and ambiguous pedestrian annotations are removed or ignored.
		The remaining valid pedestrian-related annotations are merged into a single pedestrian class for training and evaluation.
		
		\begin{table*}[t]
			\centering
			\caption{Baselines Used In The Experiments}
			\label{tab:baseline_placeholder}
			\footnotesize
			\begin{tabular}{cccc}
				\Xhline{1.2pt}
				Category & ID & Description & Selected \\
				\hline
				Registration & R1 & No-registration baseline & NoReg \\
				Registration & R2 & Template-matching-based visible--thermal registration baseline & TWMM (2022)~\cite{TWMM2022} \\
				Registration & R3 & Plug-and-play general multi-modal image registration baseline & PGMR (2025)~\cite{PGMR2025} \\
				Fusion & F1 & Detection-driven fusion baseline & DetFusion (2022)~\cite{p4_detfusion2022} \\
				Fusion & F2 & Meta-feature embedding guided infrared--visible fusion baseline & MetaFusion (2023)~\cite{p4_metafusion2023} \\
				Fusion & F3 & Registration--fusion integrated infrared--visible fusion baseline & MulFS-CAP (2025)~\cite{p2_mulfscap2025} \\
				System & S1 & Visible-only detection baseline with an external detector & YOLO11s \\
				System & S2 & Infrared-only detection baseline with an external detector & YOLO11s \\
				System & S3 & External registration + registered-IR detection pipeline & PGMR (2025)~\cite{PGMR2025} + YOLO11s \\
				System & S4 & External fusion + detection pipeline & LRRNet (2023)~\cite{LRRNet2023} + YOLO11s \\
				System & S5 & Fully external sequential Reg$\rightarrow$Fus$\rightarrow$Det pipeline & PGMR (2025)~\cite{PGMR2025} + LRRNet (2023)~\cite{LRRNet2023} + YOLO11s \\
				System & S6 & External registration preprocessing + fusion--detection integrated method & PGMR (2025)~\cite{PGMR2025} + MetaFusion (2023)~\cite{p4_metafusion2023} \\
				System & S7 & Registration--fusion integrated method + external detector & MulFS-CAP (2025)~\cite{p2_mulfscap2025} + YOLO11s \\
				\Xhline{1.2pt}
			\end{tabular}
		\end{table*}
		
		\paragraph{Patch-Based Data Preparation}
		Considering the memory cost of joint registration, fusion, and detection training, we adopt patch-based processing instead of full-frame training.
		Each visible--thermal pair is cropped into $256\times256$ patches with a stride of $128$, and at most $50$ ground-truth boxes are retained for each patch.
		The processed training patches are further split into training and validation subsets with a ratio of $0.8$.
		For background-only patches with empty samples, only $20\%$ are retained independently within the training/validation and testing splits to avoid overwhelming the optimization.
		For each retained patch, both reference-coordinate boxes and perturbed infrared-domain boxes are stored for branch-specific detection evaluation.

		\paragraph{Teacher Fusion Target}
		Since KAIST does not provide real fusion ground truth, we use the fusion results produced by LRRNet~\cite{LRRNet2023} on the original KAIST data as a pseudo teacher target for training FusDM.
		The teacher output is used only to stabilize residual fusion learning and is not treated as exact fusion ground truth for final evaluation.
		
		\paragraph{Misalignment Setting} To simulate weakly misaligned RGB-T perception, synthetic affine perturbations are applied to the infrared modality while the visible image is kept as the reference. In the current setting, the infrared patch is perturbed by an in-plane rotation up to $2^\circ$ and a translation ratio of $0.05$, corresponding to a maximum translation of approximately $12.8$ pixels along each axis for a $256\times256$ patch. This controlled setting focuses on weak affine misalignment rather than severe non-rigid deformation, and is used to evaluate whether RegDM can provide reliable geometric correction for the target RGB-T perception scenario.

		\paragraph{Evaluation Metrics}
		We evaluate TriDifNet from three perspectives.
		For registration, we report root mean squared error (RMSE), normalized cross-correlation / correlation coefficient (NCC/CC), and structural similarity (SSIM)~\cite{SSIM} between the registered infrared image and the aligned infrared reference.
		For fusion, we use weak-reference image-level measures, including CC and UIQI~\cite{UIQI}, to evaluate structural consistency and cross-modal information preservation.
		For detection, we report Precision, Recall, mAP$_{50}$, and mAP$_{50:95}$ for the visible branch, the infrared or registered-infrared branch, and the fusion branch.
		Since all valid pedestrian-related annotations are merged into a single pedestrian class, mAP$_{50:95}$ is computed in a single-class COCO-style~\cite{COCO} manner over IoU thresholds from 0.50 to 0.95 with a step size of 0.05.
		The annotation domain used for each detection branch follows the comparison setting summarized in Table~\ref{tab:comparison_protocols}. The final fusion-side detection result serves as the primary evaluation of the complete network.

		\subsection{Experimental Settings and Baselines}
		
		\paragraph{Training and Inference Settings}
		TriDifNet is implemented in PyTorch and trained on a single NVIDIA RTX 5060 Ti GPU with 16 GB memory.
		All experiments are conducted with $256\times256$ input patches.
		The model is trained under the stage-wise three-task optimization strategy introduced in Sec.~\ref{sec:training_strategy}, with a batch size of $1$, $30{,}000$ training iterations, and $500$ diffusion timesteps.
		RegDM, FusDM, and DetDM are optimized within the unified closed-loop training pipeline using branch-specific optimizers, with learning rates of $2\times10^{-4}$, $1\times10^{-4}$, and $2.5\times10^{-5}$, respectively.
		
		At test time, the training-only auxiliary components are removed, and the deployed inference pipeline consists of RegDM, FusDM, and DetDM with ROI-guided re-inference.
		For diffusion inference, RegDM and FusDM use DDIM sampling~\cite{DDIM} with $25$ steps.
		For ROI-guided re-inference, the first-pass D3 confidence threshold is set to $0.10$ to improve recall for ROI construction, whereas the final second-pass detection uses a confidence threshold of $0.30$ for reporting Precision and Recall.
		No ground-truth boxes are used for ROI construction during inference.
		All FPS results are measured on the same platform with batch size $1$, excluding data loading and file I/O.

		\paragraph{Baseline Categories}
		To avoid ambiguity, the compared methods are grouped into registration, fusion, and system-level baselines, as summarized in Table~\ref{tab:baseline_placeholder}.
		Registration baselines evaluate geometric correction under the same misaligned RGB-T input.
		Fusion baselines evaluate image-level fusion behavior under aligned-input conditions.
		System-level baselines evaluate the final pedestrian detection performance under complete RGB-T perception pipelines.
		Since LRRNet~\cite{LRRNet2023} is used as the pseudo teacher for FusDM training, it is not included as a direct image-level fusion baseline, and its downstream influence is evaluated through S4 and S5.
		For detector-based system baselines, most external pipelines use YOLO11s as the downstream detector, while S6 follows the original MetaFusion evaluation setup and uses YOLOv5s~\cite{p4_metafusion2023}.

		\paragraph{Comparison Settings} Different experiments follow different input settings. Registration comparisons start from the same misaligned RGB-T input and evaluate only the registered infrared output. Fusion comparisons are conducted under a unified aligned-input condition to isolate fusion behavior from registration errors. End-to-end comparisons start from the same misaligned RGB-T input and evaluate the final detection results with branch-specific annotations. Specifically, visible, registered-infrared, and fusion-side outputs are evaluated with reference-coordinate boxes, while unaligned-infrared detection is evaluated with perturbed infrared-domain boxes.
		
		\paragraph{Training Settings for Baselines} For fair comparison, all baseline methods are trained and evaluated under the same overall data setting used by TriDifNet. For methods with official implementations, we follow their released configurations whenever applicable, initialize them from the official pretrained weights, and fine-tune them on the current training data rather than training from scratch. For module-level baselines, registration methods are fine-tuned on the misaligned RGB-T training set, while fusion methods are fine-tuned on the aligned RGB-T training set according to Table~\ref{tab:comparison_protocols}. Each single-task baseline is fine-tuned for $50$ epochs and evaluated on the corresponding unified test set.
		
		For system-level baselines with sequential components, each downstream component is fine-tuned on the training-set outputs produced by its upstream component for $50$ epochs before being evaluated on the corresponding test-set outputs.
		For the MulFS-CAP-based~\cite{p2_mulfscap2025} baselines F3 and S7, we follow the official implementation and fine-tune it for $200$ epochs using $500$ randomly sampled training pairs and $100$ validation pairs.
		All baselines, including MulFS-CAP-based pipelines, are evaluated on the same unified test set.

		\subsection{Main Quantitative Results}
		\label{sec:main_quantitative_results}
		\paragraph{Registration Comparison} We first compare RegDM with the registration baselines under the same misalignment setting. All methods take the same misaligned RGB-T pair as input and output a registered infrared image. Since the current benchmark focuses on controlled weak affine perturbations, this comparison is intended to evaluate whether RegDM can provide reliable geometric correction under the target weakly misaligned RGB-T setting, rather than to claim general superiority under severe non-rigid deformation scenarios.

\begin{figure}[t]
			\centering
			\setlength{\tabcolsep}{0.6pt}
			\renewcommand{\arraystretch}{0.90}
			\tiny
			\begin{tabular}{cccc}
				\textbf{R1} & \textbf{R2} & \textbf{R3} & \textbf{RegDM} \\
				
				\includegraphics[width=0.235\columnwidth]{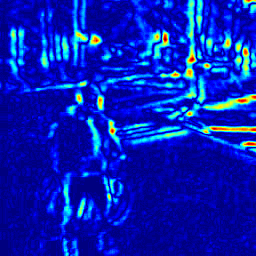} &
				\includegraphics[width=0.235\columnwidth]{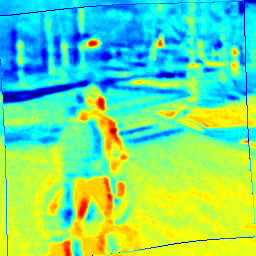} &
				\includegraphics[width=0.235\columnwidth]{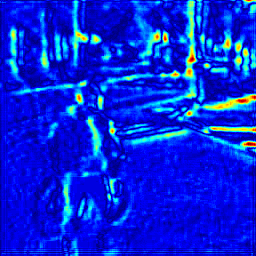} &
				\includegraphics[width=0.235\columnwidth]{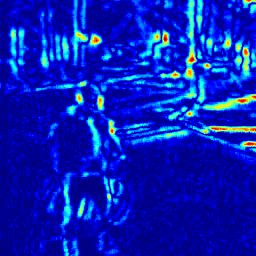} \\[-1pt]
				
				\includegraphics[width=0.235\columnwidth]{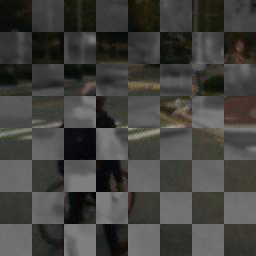} &
				\includegraphics[width=0.235\columnwidth]{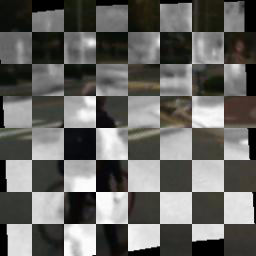} &
				\includegraphics[width=0.235\columnwidth]{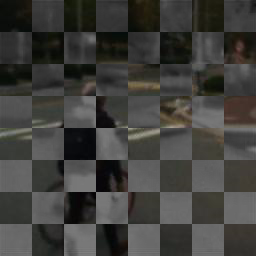} &
				\includegraphics[width=0.235\columnwidth]{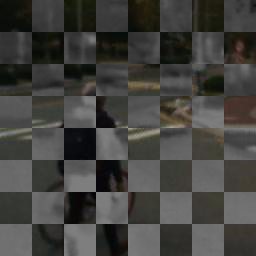} \\[1pt]
				
				\includegraphics[width=0.235\columnwidth]{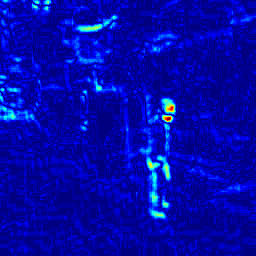} &
				\includegraphics[width=0.235\columnwidth]{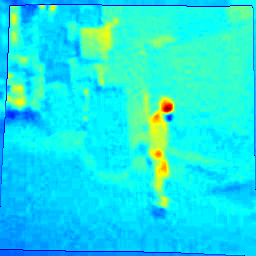} &
				\includegraphics[width=0.235\columnwidth]{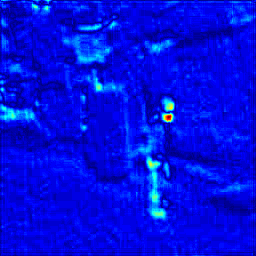} &
				\includegraphics[width=0.235\columnwidth]{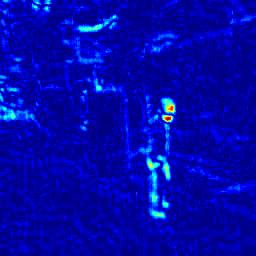} \\[-1pt]
				
				\includegraphics[width=0.235\columnwidth]{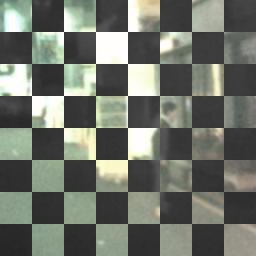} &
				\includegraphics[width=0.235\columnwidth]{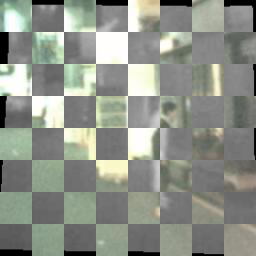} &
				\includegraphics[width=0.235\columnwidth]{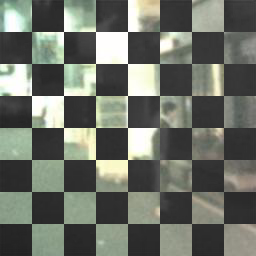} &
				\includegraphics[width=0.235\columnwidth]{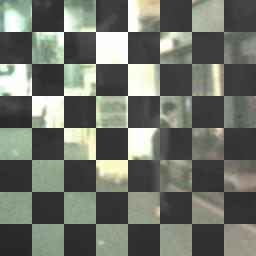}
			\end{tabular}
			
			\caption{Qualitative comparison of registration results corresponding to Table~\ref{tab:reg_comparison}, including absolute-error heatmaps and checkerboard alignment visualizations.}
			\label{fig:table3_registration_visualization}
		\end{figure}

		As shown in Table~\ref{tab:reg_comparison}, the proposed RegDM achieves the lowest RMSE (10.425901) and the highest SSIM (0.856359) among the compared registration methods, while maintaining competitive NCC/CC.
		R3 obtains a slightly higher NCC/CC (0.835497), indicating that the image-level registration scores of different methods can be close under the controlled weak-misalignment setting.
		These results suggest that RegDM provides stable geometric correction for the subsequent fusion and detection stages, rather than an overwhelming single-task registration advantage over all baselines.
		The qualitative results in Fig.~\ref{fig:table3_registration_visualization} further show that RegDM produces low-residual responses and coherent checkerboard alignment patterns.
		The downstream contribution of RegDM is further evaluated in the end-to-end detection comparison and the system-level ablations, where RegDM is coupled with Det2Reg and Fus2Reg.

		\paragraph{Fusion Comparison}
		We next compare FusDM with the fusion baselines under a unified aligned-input condition.
		This setting isolates fusion behavior from registration errors and follows the corresponding setting in Table~\ref{tab:comparison_protocols}.
		Since KAIST does not provide real fusion ground truth and the LRRNet output is used only as a pseudo teacher during training, the following comparison reports weak-reference image-level indicators rather than treating teacher-referenced scores as the primary evidence.

		\begin{table}[t]
			\centering
			\caption{Quantitative Comparison Of Registration Results}
			\label{tab:reg_comparison}
			\begin{tabular}{cccc}
				\Xhline{1.2pt}
				Method & RMSE$\downarrow$ & NCC/CC$\uparrow$ & SSIM$\uparrow$ \\
				\hline
				R1 & 10.976204 & 0.831235 & 0.854959 \\
				R2 & 45.652067 & 0.591024 & 0.543002 \\
				R3 & 10.457858 & \textbf{0.835497} & 0.853929 \\
				Proposed RegDM & \textbf{10.425901} & 0.832831 & \textbf{0.856359} \\
				\Xhline{1.2pt}
			\end{tabular}
		\end{table}
		
		\afterpage{\begin{figure*}[!b]
	\centering
	\setlength{\tabcolsep}{0.8pt}
	\renewcommand{\arraystretch}{0.88}
	\scriptsize
	
	\newcommand{\detimg}[1]{\raisebox{-0.5\height}{\includegraphics[width=0.085\textwidth]{#1}}}
	\newcommand{\casecell}[1]{\raisebox{-0.5\height}{\textbf{#1}}}
	
	\begin{tabular}{@{}ccccccccccc@{}}
		& \textbf{VIS} & \textbf{Unaligned IR} & \textbf{S1} & \textbf{S2} & \textbf{S3} & \textbf{S4} & \textbf{S5} & \textbf{S6} & \textbf{S7} & \textbf{TriDifNet} \\
		
		\casecell{Case 1} &
		\detimg{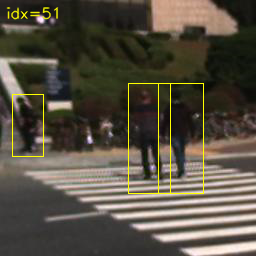} &
		\detimg{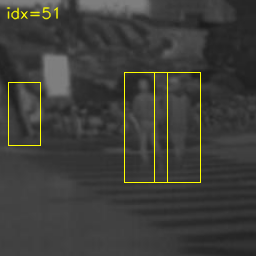} &
		\detimg{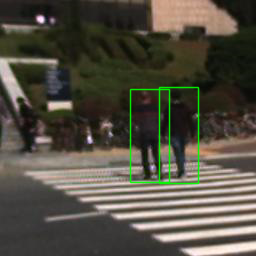} &
		\detimg{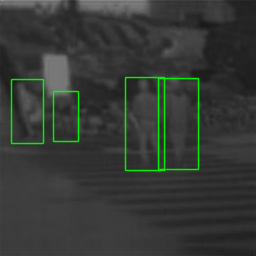} &
		\detimg{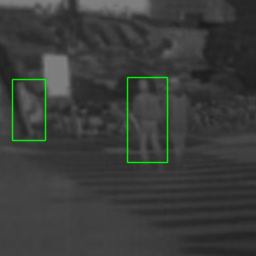} &
		\detimg{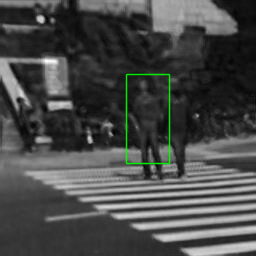} &
		\detimg{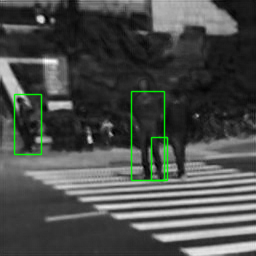} &
		\detimg{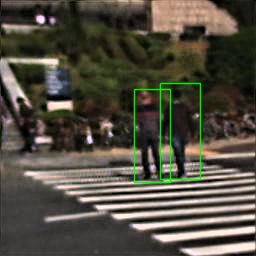} &
		\detimg{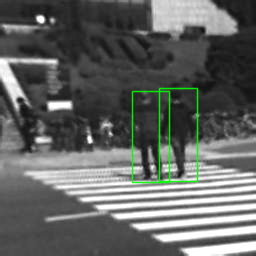} &
		\detimg{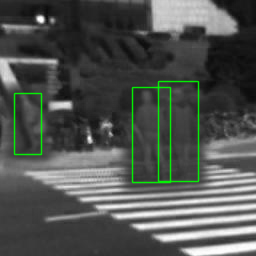} \\[-2pt]
		
		\casecell{Case 2} &
		\detimg{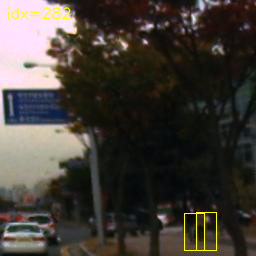} &
		\detimg{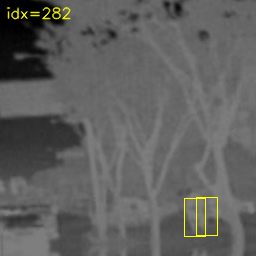} &
		\detimg{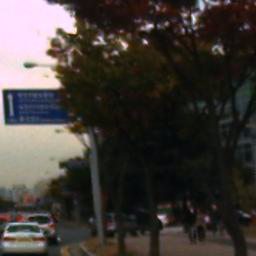} &
		\detimg{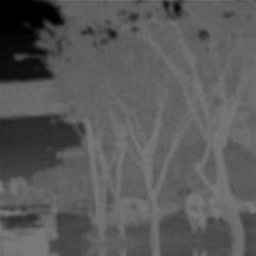} &
		\detimg{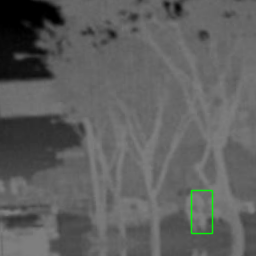} &
		\detimg{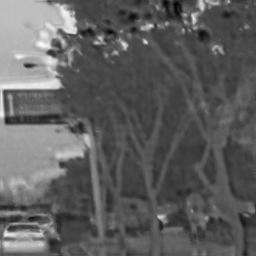} &
		\detimg{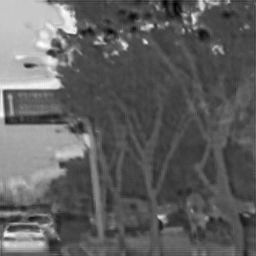} &
		\detimg{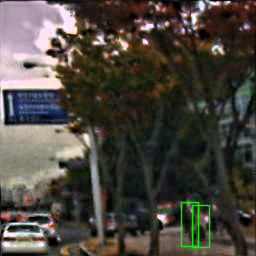} &
		\detimg{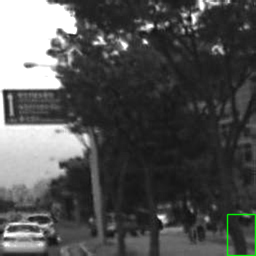} &
		\detimg{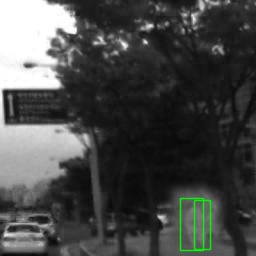} \\[-2pt]
		
		\casecell{Case 3} &
		\detimg{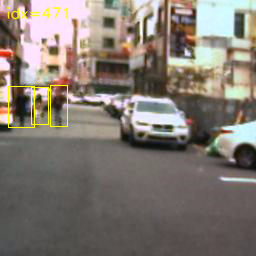} &
		\detimg{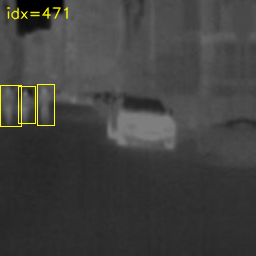} &
		\detimg{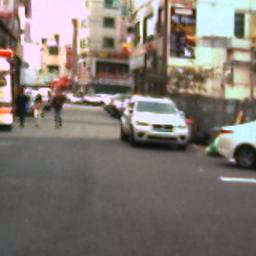} &
		\detimg{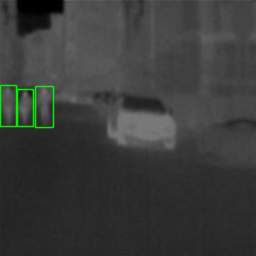} &
		\detimg{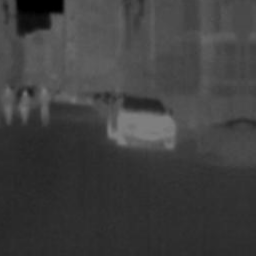} &
		\detimg{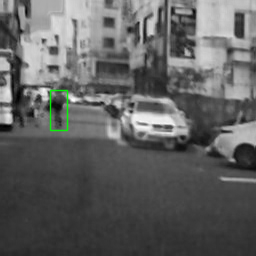} &
		\detimg{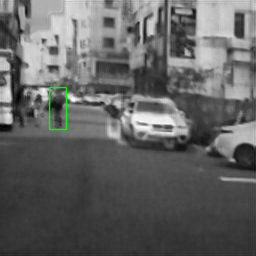} &
		\detimg{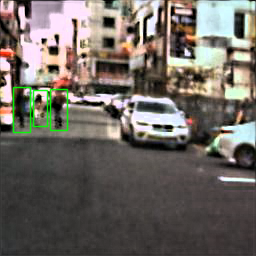} &
		\detimg{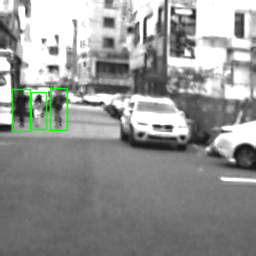} &
		\detimg{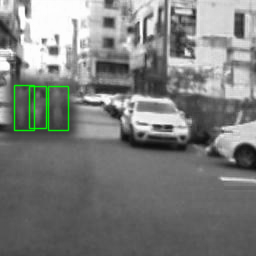} \\[-2pt]
		
		\casecell{Case 4} &
		\detimg{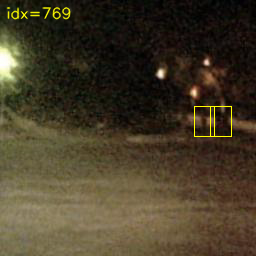} &
		\detimg{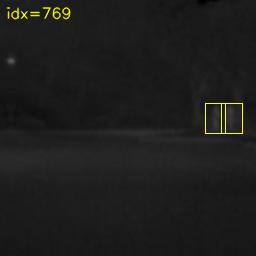} &
		\detimg{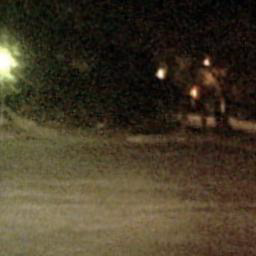} &
		\detimg{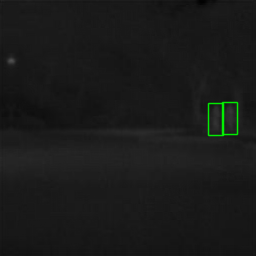} &
		\detimg{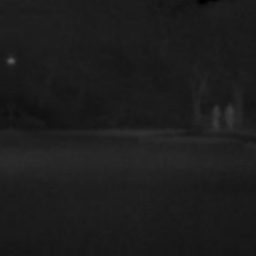} &
		\detimg{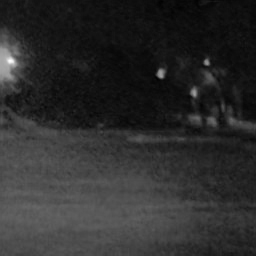} &
		\detimg{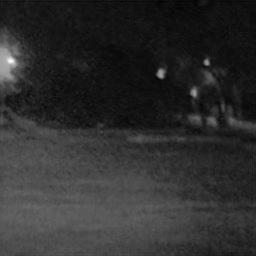} &
		\detimg{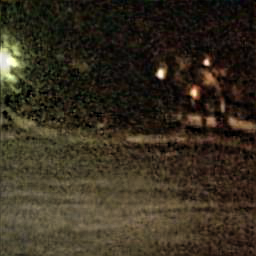} &
		\detimg{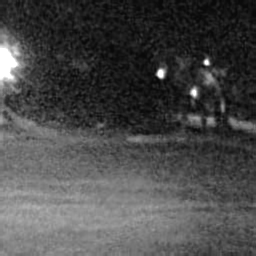} &
		\detimg{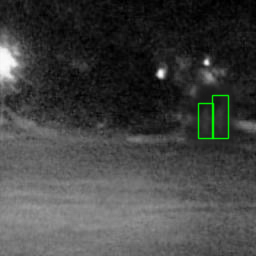} \\[-2pt]
		
		\casecell{Case 5} &
		\detimg{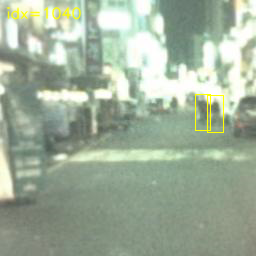} &
		\detimg{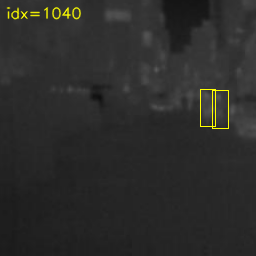} &
		\detimg{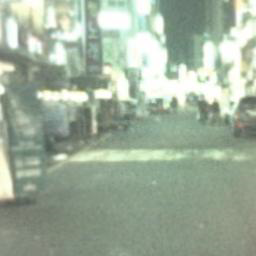} &
		\detimg{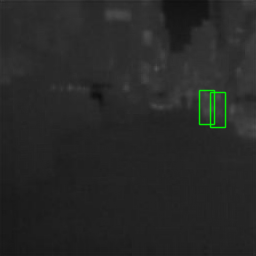} &
		\detimg{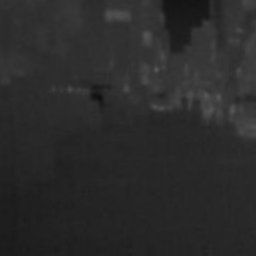} &
		\detimg{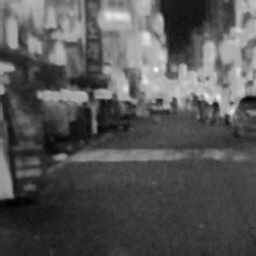} &
		\detimg{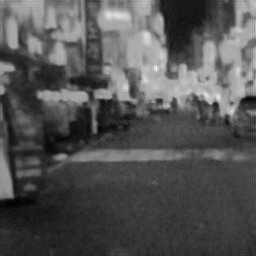} &
		\detimg{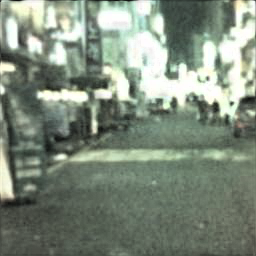} &
		\detimg{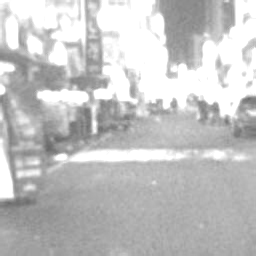} &
		\detimg{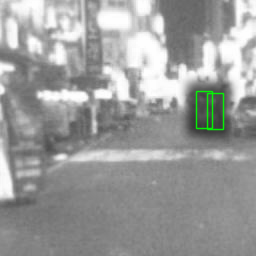}
	\end{tabular}
	
	\caption{Qualitative comparison of end-to-end pedestrian detection results corresponding to Table~\ref{tab:e2e_detection_comparison}. 
		Case~1 shows daytime large targets; Cases~2 and~3 show daytime small targets; Case~4 shows nighttime small targets under low-light conditions; and Case~5 shows nighttime small targets with relatively stronger illumination. 
		Yellow and green boxes denote ground truth and predictions, respectively.}
	\label{fig:table5_detection_visualization}
\end{figure*}}
		
		\begin{table}[t]
			\centering
			\caption{Quantitative Comparison Of Fusion Results}
			\label{tab:fusion_comparison}
			\begin{tabular}{ccc}
				\Xhline{1.2pt}
				Method & CC$\uparrow$ & UIQI$\uparrow$ \\
				\hline
				F1 & 0.900771 & 0.808711 \\
				F2 & 0.903327 & \textbf{0.864656} \\
				F3 & \textbf{0.942980} & 0.862271 \\
				Proposed FusDM & 0.926469 & 0.840299 \\
				\Xhline{1.2pt}
			\end{tabular}
		\end{table}
		
		\begin{figure}[t]
			\centering
			\setlength{\tabcolsep}{0.4pt}
			\renewcommand{\arraystretch}{0.88}
			\tiny
			\begin{tabular}{ccccc}
				\textbf{F1} & \textbf{F2} & \textbf{F3} & \textbf{FusDM} & \textbf{FusDM ROI-IR injection} \\
				
				\includegraphics[width=0.188\columnwidth]{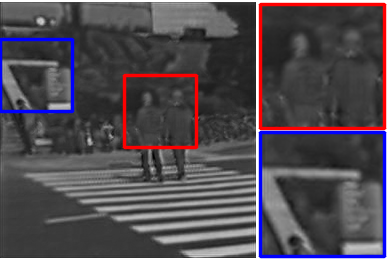} &
				\includegraphics[width=0.188\columnwidth]{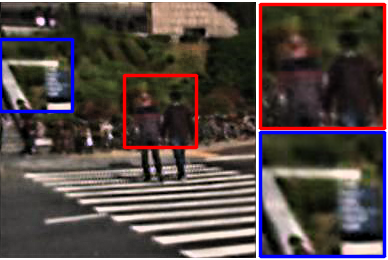} &
				\includegraphics[width=0.188\columnwidth]{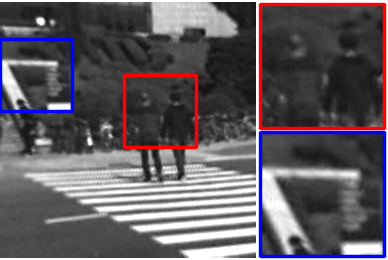} &
				\includegraphics[width=0.188\columnwidth]{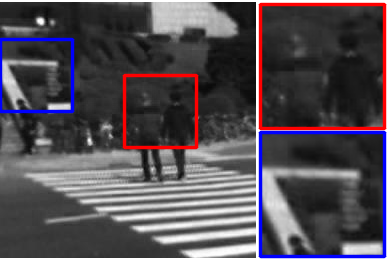} &
				\includegraphics[width=0.188\columnwidth]{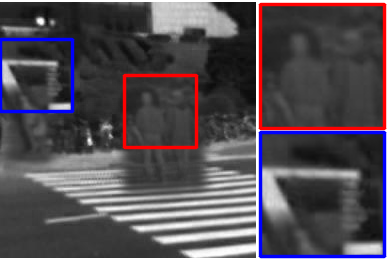} \\[-1pt]
				
				\includegraphics[width=0.188\columnwidth]{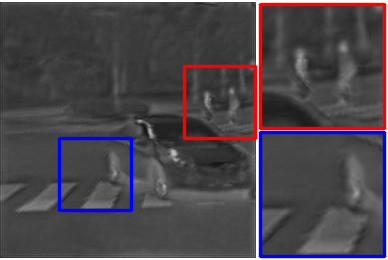} &
				\includegraphics[width=0.188\columnwidth]{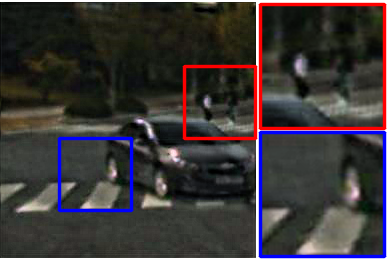} &
				\includegraphics[width=0.188\columnwidth]{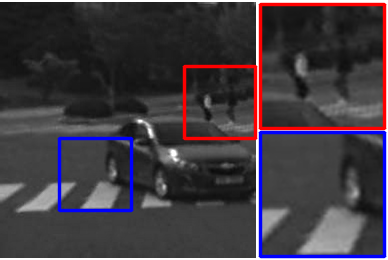} &
				\includegraphics[width=0.188\columnwidth]{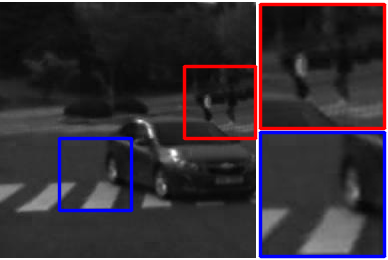} &
				\includegraphics[width=0.188\columnwidth]{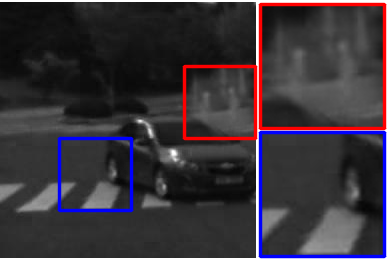} \\[-1pt]
				
				\includegraphics[width=0.188\columnwidth]{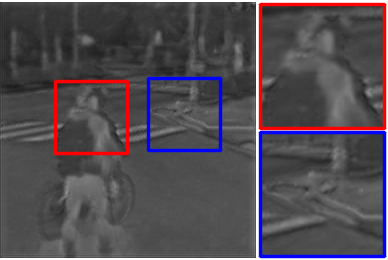} &
				\includegraphics[width=0.188\columnwidth]{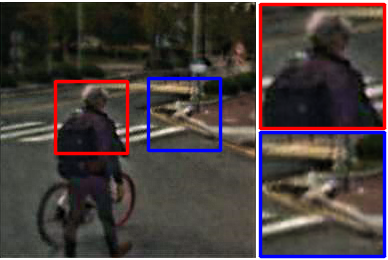} &
				\includegraphics[width=0.188\columnwidth]{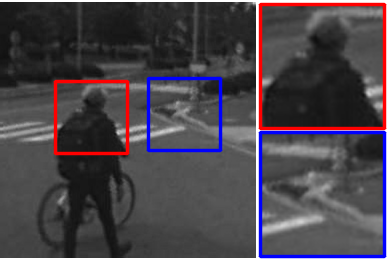} &
				\includegraphics[width=0.188\columnwidth]{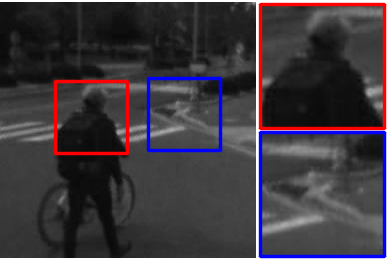} &
				\includegraphics[width=0.188\columnwidth]{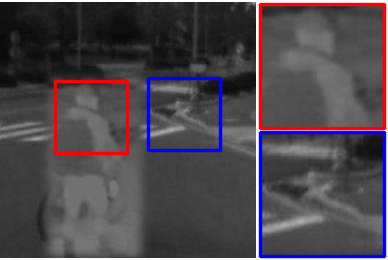} \\[-1pt]
				
				\includegraphics[width=0.188\columnwidth]{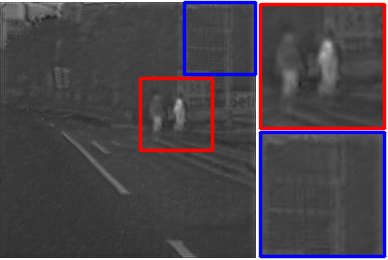} &
				\includegraphics[width=0.188\columnwidth]{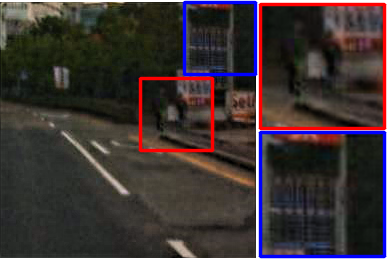} &
				\includegraphics[width=0.188\columnwidth]{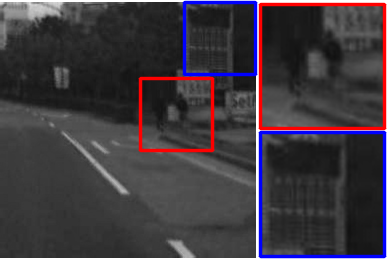} &
				\includegraphics[width=0.188\columnwidth]{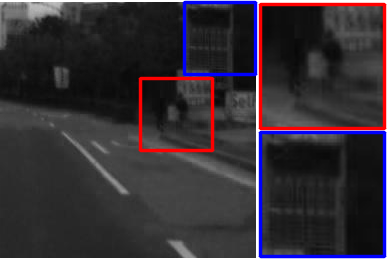} &
				\includegraphics[width=0.188\columnwidth]{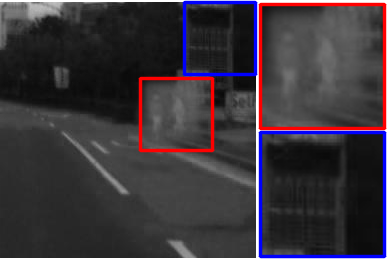}
			\end{tabular}
			\caption{Qualitative comparison of fusion results corresponding to Table~\ref{tab:fusion_comparison}. TriDifNet emphasizes target-aware infrared saliency while maintaining visible structural cues for downstream detection.}
			\label{fig:table4_fusion_visualization}
		\end{figure}

		Table~\ref{tab:fusion_comparison} reports weak-reference image-level fusion scores under the aligned-input setting.
		F3 and F2 achieve the best CC and UIQI, respectively, while FusDM obtains competitive but not leading teacher-referenced scores.
		Since the KAIST benchmark lacks real fusion ground truth, these metrics are used only to characterize image-level fusion behavior rather than downstream perception utility.
		Fig.~\ref{fig:table4_fusion_visualization} shows that FusDM preserves visible structural cues and enhances target-related infrared responses.
		More detailed evidence for its task-level effectiveness is provided by the end-to-end detection comparison in Sec.~\ref{sec:main_quantitative_results} and the cross-task ablations in Sec.~\ref{sec:ablation_studies}.

		\paragraph{End-to-End Detection Comparison}
		The most important comparison is the end-to-end pedestrian detection performance.
		All system-level baselines start from the same misaligned RGB-T input and are evaluated under the same detection benchmark.
		This comparison directly verifies whether the proposed closed-loop Reg--Fus--Det optimization improves downstream perception compared with single-modality settings, sequential external pipelines, and partially coupled alternatives.

		\begin{table}[t]
			\centering
			\caption{End-To-End Pedestrian Detection Comparison}
			\label{tab:e2e_detection_comparison}
			\footnotesize
			\setlength{\tabcolsep}{4.5pt}
			\renewcommand{\arraystretch}{0.95}
			\begin{tabular}{ccccc}
				\Xhline{1.2pt}
				Method & P(\%)$\uparrow$ & R(\%)$\uparrow$ & mAP$_{50}$(\%)$\uparrow$ & mAP$_{50:95}$(\%)$\uparrow$ \\
				\hline
				S1 & \textbf{81.53} & 39.49 & 36.02 & 16.45 \\
				S2 & 79.39 & 44.80 & 40.33 & 20.88 \\
				S3 & 81.14 & 9.28 & 8.32 & 3.12 \\
				S4 & 80.22 & 25.22 & 22.90 & 10.15 \\
				S5 & 69.53 & 37.71 & 32.65 & 14.66 \\
				S6 & 41.62 & 52.84 & 38.51 & 18.76 \\
				S7 & 72.42 & 48.45 & 43.66 & 19.02 \\
				Proposed TriDifNet & 73.85 & \textbf{65.44} & \textbf{57.55} & \textbf{26.83} \\
				\Xhline{1.2pt}
			\end{tabular}
		\end{table}

		Table~\ref{tab:e2e_detection_comparison} reports the system-level detection results.
		TriDifNet obtains the best scores on the three main detection metrics, i.e., 65.44\% Recall, 57.55\% mAP$_{50}$, and 26.83\% mAP$_{50:95}$.
		Compared with the strongest external system-level baseline S7, TriDifNet improves these three metrics by 16.99\%, 13.89\%, and 7.81\%, respectively.
		Although the visible-only baseline S1 obtains the highest precision, its recall and mAP remain limited, indicating that single-modality detection is insufficient under misaligned RGB-T inputs.
		
		S3 shows a sharp degradation under the post-registration detection setting, with only 9.28\% recall, 8.32\% mAP$_{50}$, and 3.12\% mAP$_{50:95}$.
		Although PGMR obtains competitive image-level registration scores in Table~\ref{tab:reg_comparison}, its registered output may still contain object-level residual offsets with respect to the reference-domain detection annotations.
		Such offsets can be amplified into localization errors and missed detections, leading to a clear decrease in recall and mAP.
		This result shows that good single-task registration metrics do not necessarily guarantee detection-consistent alignment in a decoupled Reg$\rightarrow$Det pipeline.
		
		S4 also yields limited detection performance, with 25.22\% recall and 22.90\% mAP$_{50}$.
		Since LRRNet is applied without explicit registration in this setting, the fused image can suffer from misalignment-induced ghosting, double edges, and target deformation.
		These artifacts degrade the detector input and further propagate to the final detection metrics.
		S5 performs the fully external Reg$\rightarrow$Fus$\rightarrow$Det pipeline, but the sequential combination still suffers from the accumulation of registration and fusion errors.
		For the MetaFusion-based setting S6, we follow its original evaluation setup with YOLOv5s, so part of the performance difference may also come from the different detector configuration.
		
		Overall, the large gains in recall and mAP indicate that TriDifNet reduces missed detections and achieves a better detection trade-off under misaligned RGB-T inputs.
		These results support the effectiveness of closed-loop optimization among registration, fusion, and detection over sequential or partially coupled pipelines.
		
		Fig.~\ref{fig:table5_detection_visualization} further provides qualitative evidence for the end-to-end comparison.
		Compared with the system-level baselines, TriDifNet produces more complete and better localized pedestrian predictions under misaligned RGB-T inputs.
		This advantage is most evident in Cases~4 and~5, where several baselines suffer from missed detections under weak visible cues and low-contrast infrared responses, whereas TriDifNet still retains stable target predictions.
		The qualitative comparison is consistent with the quantitative gains reported in Table~\ref{tab:e2e_detection_comparison}.
		
		\begin{table}[t]
			\caption{System-Level Ablation Study}
			\label{tab:system_ablation}
			\centering
			\footnotesize
			\setlength{\tabcolsep}{3.5pt}
			\renewcommand{\arraystretch}{1.08}
			\begin{tabular}{cc|ccccc}
				\Xhline{1.2pt}
				\multicolumn{2}{c|}{Method} 
				& \multirow{2}{*}{P(\%)$\uparrow$} 
				& \multirow{2}{*}{R(\%)$\uparrow$} 
				& \multirow{2}{*}{mAP$_{50}$(\%)$\uparrow$} 
				& \multirow{2}{*}{mAP$_{50:95}$(\%)$\uparrow$}
				& \multirow{2}{*}{FPS$\uparrow$} \\
				RegDM & FusDM &  &  &  &  &  \\
				\hline
				-- & -- & 48.77 & 49.57 & 37.90 & 13.71 & \textbf{11.02} \\
				-- & \checkmark & 70.81 & 57.34 & 44.67 & 19.13 & 0.32 \\
				\checkmark & -- & 51.03 & 49.46 & 39.04 & 14.07 & 1.03 \\
				\checkmark & \checkmark & \textbf{73.85} & \textbf{65.44} & \textbf{57.55} & \textbf{26.83} & 0.25 \\
				\Xhline{1.2pt}
			\end{tabular}
		\end{table}
		
		\begin{table*}[!b]
			\centering
			\caption{Ablation Study On Cross-Task Feedback}
			\label{tab:cross_task_feedback_ablation}
			\renewcommand{\arraystretch}{1.10}
			\setlength{\tabcolsep}{3.2pt}
			\scriptsize
			\begin{tabular*}{\textwidth}{@{\extracolsep{\fill}}c c c c c c|c c c|c c|c c c c@{}}
				\Xhline{1.2pt}
				\multirow{2}{*}{Index}
				& \multirow{2}{*}{Det2Reg}
				& \multirow{2}{*}{Fus2Reg}
				& \multirow{2}{*}{Det2Fus}
				& \multirow{2}{*}{ROI-IR Injection}
				& \multirow{2}{*}{Nash-MTL}
				& \multicolumn{3}{c|}{Registration}
				& \multicolumn{2}{c|}{Fusion}
				& \multicolumn{4}{c}{Detection} \\
				& & & & & 
				& RMSE$\downarrow$
				& NCC/CC$\uparrow$
				& SSIM$\uparrow$
				& CC$\uparrow$
				& UIQI$\uparrow$
				& P(\%)$\uparrow$
				& R(\%)$\uparrow$
				& mAP$_{50}$(\%)$\uparrow$
				& mAP$_{50:95}$(\%)$\uparrow$ \\
				\midrule
				
				1 & -- & \checkmark & \checkmark & \checkmark & \checkmark
				& 10.8720 & 0.8215 & 0.8461
				& 0.9258 & 0.8447
				& 71.30 & 62.77 & 54.11 & 24.73 \\
				
				2 & \checkmark & -- & \checkmark & \checkmark & \checkmark
				& 11.0282 & 0.8176 & 0.8426
				& 0.9263 & 0.8427
				& 73.80 & 61.48 & 55.59 & 26.75 \\
				
				3 & \checkmark & \checkmark & -- & \checkmark & \checkmark
				& 10.9231 & 0.8188 & 0.8448
				& \textbf{0.9665} & \textbf{0.9492}
				& 55.17 & 52.17 & 42.03 & 18.70 \\
				
				4 & \checkmark & \checkmark & -- & -- & \checkmark
				& 11.0814 & 0.8181 & 0.8424
				& 0.9640 & 0.9406
				& 54.92 & 47.01 & 36.60 & 14.54 \\
				
				5 & \checkmark & \checkmark & \checkmark & \checkmark & --
				& 10.9469 & 0.8196 & 0.8443
				& 0.9312 & 0.8508
				& 71.56 & 63.59 & 55.32 & 25.52 \\
				
				6 & \checkmark & \checkmark & \checkmark & \checkmark & \checkmark
				& \textbf{10.4259} & \textbf{0.8328} & \textbf{0.8563}
				& 0.9264 & 0.8402
				& \textbf{73.85} & \textbf{65.44} & \textbf{57.55} & \textbf{26.83} \\
				
				\Xhline{1.2pt}
			\end{tabular*}
		\end{table*}
		
		\subsection{Ablation Studies}
		\label{sec:ablation_studies}
		
		\paragraph{System-Level Ablation}
		We conduct system-level ablations to evaluate the contribution of RegDM and FusDM to RGB-T pedestrian perception.
		DetDM is retained in all variants, and each variant is trained under the same training schedule before testing.
		When both RegDM and FusDM are removed, detection is performed on the unaligned infrared input; when FusDM is removed, detection is performed on the registered infrared image; otherwise, detection is performed on the fusion-oriented representation.
		
		As shown in Table~\ref{tab:system_ablation}, removing RegDM and FusDM gives limited detection performance, with 37.90\% mAP$_{50}$ and 13.71\% mAP$_{50:95}$.
		Using FusDM alone improves mAP$_{50}$ from 37.90\% to 44.67\% and mAP$_{50:95}$ from 13.71\% to 19.13\%, indicating that cross-modal fusion provides useful perception cues even under misalignment.
		Using RegDM alone produces a smaller improvement, increasing mAP$_{50}$ from 37.90\% to 39.04\% and mAP$_{50:95}$ from 13.71\% to 14.07\%.
		This suggests that registration alone is insufficient for the final perception task when cross-modal fusion is absent.
		The full TriDifNet achieves the best detection results, reaching 65.44\% recall, 57.55\% mAP$_{50}$, and 26.83\% mAP$_{50:95}$.
		The large performance gap between the full TriDifNet and the single-module variants indicates that RegDM and FusDM are more effective when jointly optimized within the full TriDifNet model.
		
		Table~\ref{tab:system_ablation} also reports the inference efficiency of different variants.
		The lower FPS of the full TriDifNet is mainly associated with the additional iterative computation introduced by diffusion-based registration, diffusion-based fusion, and ROI-guided re-inference.
		This efficiency cost is a common limitation of diffusion-based registration--fusion methods~\cite{p2_rfdifnet2025}.
		In this work, the main focus is to evaluate three-task closed-loop optimization under misaligned RGB-T inputs.
		Improving the real-time efficiency of the diffusion-based inference pipeline will be investigated in future work.
		\paragraph{Cross-Task and Internal Mechanism Ablation}
		We further evaluate the cross-task feedback paths and the key internal mechanisms in Table~\ref{tab:cross_task_feedback_ablation}.
		Compared with the full TriDifNet, removing Det2Reg degrades both registration and detection, increasing RMSE from 10.4259 to 10.8720 and reducing mAP$_{50}$ from 57.55\% to 54.11\%.
		This indicates that the frozen infrared-domain detector provides useful object-level semantic guidance for registration.
		Removing Fus2Reg causes the largest registration degradation among the feedback-path variants, with RMSE increasing to 11.0282 and SSIM decreasing to 0.8426, suggesting that fusion-to-registration structural feedback contributes to geometric consistency.
		
		The detection-aware fusion pathway has the most direct influence on final perception.
		Without Det2Fus, mAP$_{50}$ and mAP$_{50:95}$ decrease from 57.55\% and 26.83\% to 42.03\% and 18.70\%, respectively.
		When ROI-IR Injection is further disabled, the performance further drops to 36.60\% and 14.54\%.
		The variants without Det2Fus or ROI-IR Injection obtain higher fusion scores in terms of CC and UIQI, but these higher image-level scores are accompanied by a clear decline in detection performance.
		This trend suggests that removing detection-oriented feedback shifts the fusion branch toward weak-reference image-level similarity, rather than target-aware representation learning.
		Therefore, weak-reference fusion metrics do not necessarily reflect downstream task utility in the misaligned RGB-T detection setting.
		The performance drops caused by removing Det2Fus and ROI-IR Injection support the importance of fusion--detection coupling and ROI-level infrared injection for constructing a task-discriminative fusion representation.
		
		Removing Nash-MTL also reduces mAP$_{50}$ from 57.55\% to 55.32\%, indicating the benefit of late-stage three-branch optimization balancing among registration, fusion, and detection.
		Overall, the full TriDifNet achieves the best registration and detection results, supporting the complementary effects of the proposed closed-loop feedback paths and internal refinement mechanisms.

		\section{Conclusion}
		In this paper, we propose TriDifNet, a unified diffusion-based network for misaligned RGB-T pedestrian perception.
		By optimizing registration, fusion, and detection in a closed loop, TriDifNet reduces sequential error propagation and enables geometric alignment, cross-modal fusion, and task-level semantics to reinforce each other.
		Experiments on the KAIST multispectral pedestrian benchmark show that TriDifNet provides reliable registration and task-oriented fusion, while achieving clear improvements in end-to-end pedestrian detection over single-modality, sequential, and partially coupled baselines.
		Ablation studies further support the effectiveness of the cross-task feedback paths, ROI-guided infrared injection, and late-stage multi-task balancing.
		Future work will focus on improving inference efficiency and extending the network to broader RGB-T perception scenarios.

		% or
		

		\ifCLASSOPTIONcaptionsoff
		\newpage
		\fi

		
		
		%\bibliographystyle{IEEEtran}
		\bibliographystyle{IEEEtran}
		\bibliography{paper}
		
		%\begin{biography}[{\includegraphics[width=1in,height=1.25in,clip,keepaspectratio]{mshell}}]{Michael Shell}
		
		\begin{IEEEbiography}[{\includegraphics[width=1in,height=1.25in,clip,keepaspectratio]{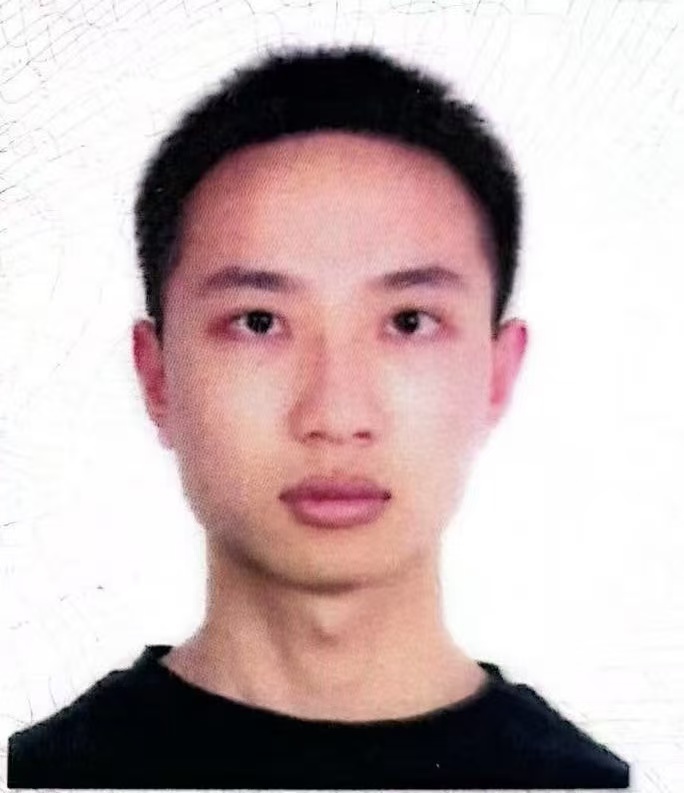}}]{Qianchao He}
			received the B.Eng. degree in aircraft control and information engineering from Northwestern Polytechnical University, Xi’an, China, in 2025. He is currently pursuing the M.Eng. degree in control science and engineering with the School of Automation Science and Engineering, Xi'an Jiaotong University, Xi'an, China.
			
			His research interests include computer vision and diffusion models. 
		\end{IEEEbiography}
		
		\begin{IEEEbiography}[{\includegraphics[width=1in,height=1.25in,clip,keepaspectratio]{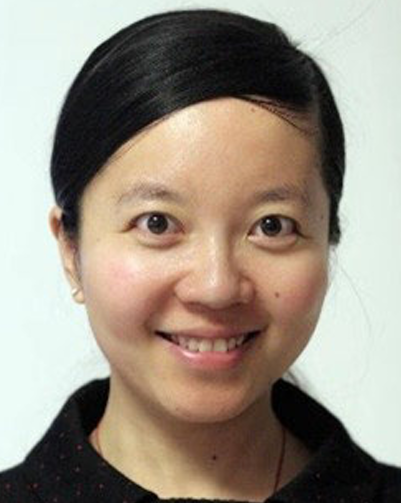}}]{Jing Liu}
			(Senior Member, IEEE)
			received the B.Sc. and M.Sc. degrees in automatic control from Northwestern Polytechnical University, Xi’an, China, in 1998 and 2000, respectively, and the Ph.D. degree in electrical and computer engineering from the National University of Singapore, Singapore, in 2007.
			
			From 2006 to 2008, she was a Research Fellow with the Department of Electrical and Computer Engineering, National University of Singapore. She is currently a Professor with the School of Electronics and Information Engineering, Xi’an Jiaotong University, Xi’an. Her research interests include tensor decomposition, information fusion, signal processing, and applications in target detection, recognition, and tracking.
		\end{IEEEbiography}
		
		\begin{IEEEbiography}[{\includegraphics[width=1in,height=1.25in,clip,keepaspectratio]{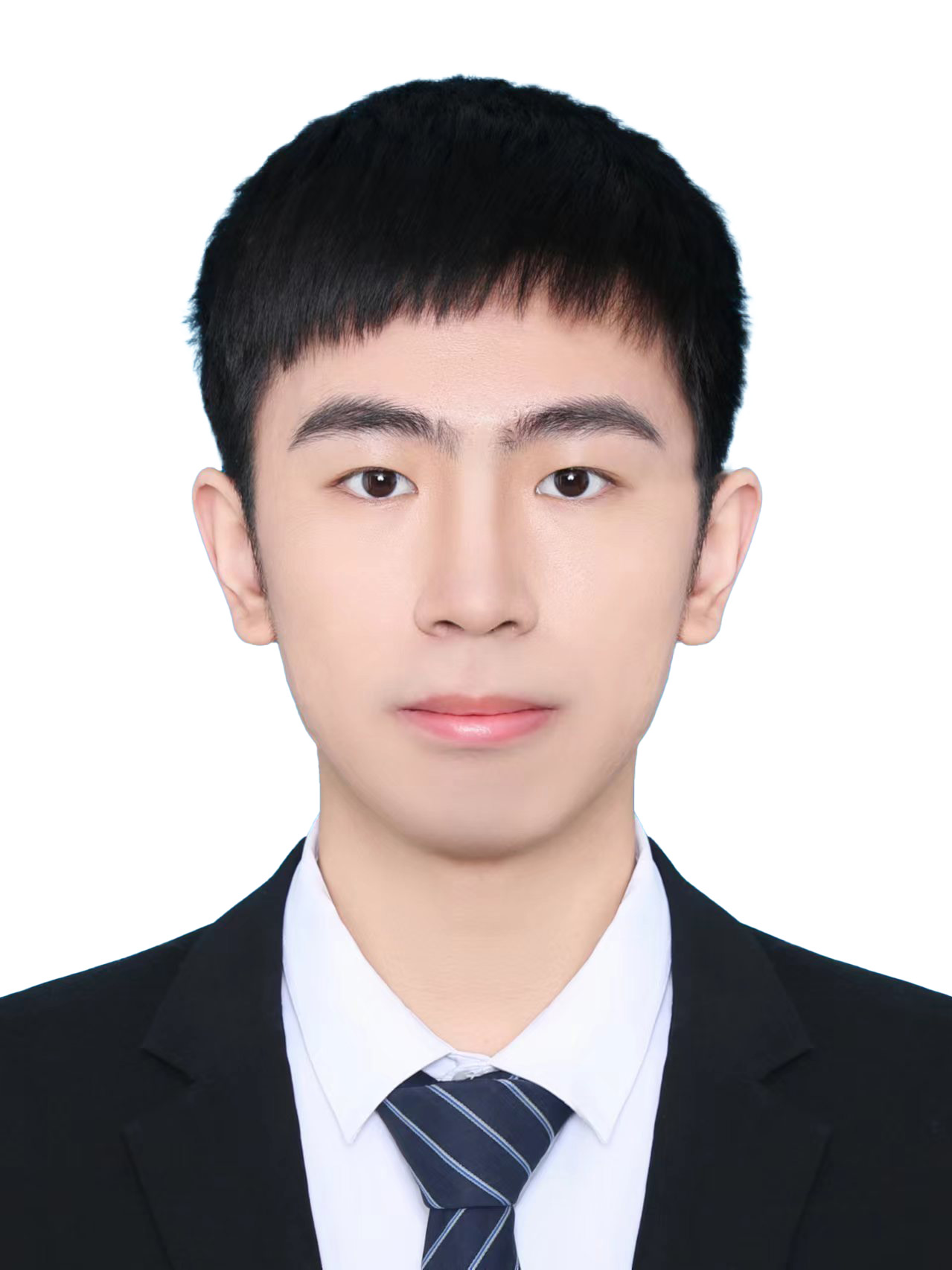}}]{Yunhao Li}
			received the B.Eng. degree in automation from Xi'an Jiaotong University, Xi’an, China, in 2025. He is currently pursuing the M.Eng. degree in control science and engineering with the School of Automation Science and Engineering, Xi'an Jiaotong University, Xi'an, China.
			
			His research interests include image processing, target tracking and deep learning. 
		\end{IEEEbiography}

		\begin{IEEEbiography}[{\includegraphics[width=1in,height=1.25in,clip,keepaspectratio]{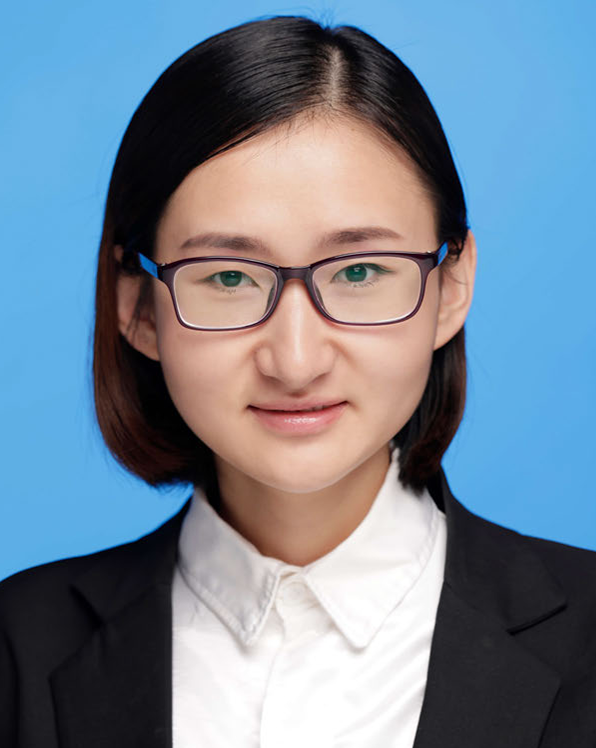}}]{Xiaoqing Tian}
			received the M.E. degree from Ningxia University, Yinchuan, China, in 2017, and the Ph.D. degree in control science and engineering from Xi’an Jiaotong University, Xi’an, China, in 2024. She is currently a Lecturer with the School of Electrical Engineering and Automation, Xi’an University of Science and Technology, Xi’an. 
			
			Her research interests include radar signal processing and target tracking.
		\end{IEEEbiography}

		
		%\vfill
		

	\end{document}